\documentclass{article}

\usepackage{arxiv}

\usepackage[utf8]{inputenc} 
\usepackage[T1]{fontenc}    

\usepackage{amssymb}
\usepackage{amsmath}
\usepackage{algorithm}
\usepackage{algorithmicx}
\usepackage{algpseudocode}
\usepackage{multirow}
\usepackage{etoolbox}
\usepackage{tabularx}
\usepackage{enumitem}
\usepackage{subcaption}

\usepackage{hyperref}       
\usepackage{url}            
\usepackage{booktabs}       
\usepackage{amsfonts}       
\usepackage{nicefrac}       
\usepackage{microtype}      
\usepackage{lipsum}
\usepackage{graphicx}

\algdef{SE}[DOWHILE]{Do}{doWhile}{\algorithmicdo}[1]{\algorithmicwhile\ #1}
\DeclareMathOperator*{\argmin}{argmin}

\title{From Node Embedding To Community Embedding : A Hyperbolic Approach}

\author{
 Thomas Gerald\thanks{Equal contributions.}\\
Science Sorbonne Université, Lip6\\
Paris, France\\
  \texttt{thomas.gerald@lip6.fr} \\
   \And
Hadi Zaatiti$^*$\\
IRT SystemX\\
Palaiseau, France\\
  \texttt{hadi.zaatiti@irt-systemx.fr} \\
  \And
Hatem Hajri$^*$\\
IRT SystemX\\
Palaiseau, France\\
  \texttt{hatem.hajri@irt-systemx.fr} \\
   \And
Nicolas Baskiotis \\
Science Sorbonne Université, Lip6\\
Paris, France\\
  \texttt{nicolas.baskiotis@lip6.fr} \\
   \And
Olivier Schwander\\
Science Sorbonne Université, Lip6 \\
Paris, France\\
\texttt{olivier.schwander@lip6.fr} \\
}

\begin{document}
\maketitle
\begin{abstract}

Detecting communities on graphs has received significant interest in recent literature. Current state-of-the-art community embedding approach called \textit{ComE} tackles this problem by coupling graph embedding with community detection. 
Considering the success of hyperbolic representations of graph-structured data in last years, an ongoing challenge is to
set up a hyperbolic approach for the community detection problem. 

The present paper meets this challenge by introducing a Riemannian equivalent of \textit{ComE}. Our proposed  approach combines hyperbolic embeddings with Riemannian K-means or Riemannian mixture models to perform community detection. We illustrate the usefulness of this framework through several experiments on real-world social networks and comparisons with \textit{ComE} and recent hyperbolic-based classification approaches.

\end{abstract}

\section{Introduction}
In recent years, the idea of embedding data in new spaces has proven effective in many applications. Indeed, several techniques have become very popular due to their great ability to represent data while reducing the complexity and dimensionality of the space. For instance, Word2vec \cite{NIPS2013_5021} and Glove \cite{Pennington14glove:global} are widely used tools in natural language processing, Nod2vec \cite{Grover2016node2vecSF}, Graph2vec \cite{DBLP:journals/corr/NarayananCVCLJ17} and DeepWalk \cite{Perozzi:2014:DOL:2623330.2623732} are commonly used for community detection, link prediction and node classification in social networks \cite{Cui2017ASO}. \\
\noindent The present paper is concerned with learning Graph Structured Data (GSD). Examples of such data include social networks, hierarchical lexical databases such as Wordnet \cite{miller1995wordnet} and Lexical entailments datasets such as Hyperlex \cite{vuli2016hyperlex}. In the state-of-the-art, one can distinguish two different approaches to cluster this type of data. The first one applies pure clustering techniques on graphs such as spectral clustering algorithms \cite{Spielmat:1996:SPW:874062.875505}, power iteration clustering \cite{Lin2010PowerIC} and label propagation \cite{Zhu02learningfrom}. The second one is two-step and may be called Euclidean clustering after (Euclidean) embedding. First it embedds data in Euclidean spaces using techniques such as Nod2vec, Graph2vec and DeepWalk and then applies traditional clustering techniques such as $K$-means algorithms. This approach appeared notably in \cite{DBLP:journals/corr/ZhengCCCC16,DBLP:journals/corr/TuWZLS16,Wang:2016:SDN:2939672.2939753}.
More recently the \textit{ComE} algorithm \cite{cavallari2017learning} achieved state-of-the-art performances in detecting communities on graphs. The main idea of this algorithm is to alternate between embedding and learning communities by means of Gaussian mixture models.

\noindent Learning GSD has known a major achievement in recent years thanks to the discovery of hyperbolic embeddings. Although it has been speculated since several years that hyperbolic spaces would better represent GSD than Euclidean spaces \cite{Gromov1987,PhysRevE,hhh,6729484}, it is only recently that these speculations have been proven effective through concrete studies and applications \cite{NIPS2017_7213,DBLP:journals/corr/ChamberlainCD17,DBLP:conf/icml/SalaSGR18}. As outlined by \cite{NIPS2017_7213}, Euclidean embeddings require large dimensions to capture certain complex relations such as the Wordnet noun hierarchy. On the other hand, this complexity can be captured by a simple model of hyperbolic geometry such as the Poincar\'e disc of two dimensions \cite{DBLP:conf/icml/SalaSGR18}. Additionally, hyperbolic embeddings provide better visualisation of clusters on graphs than Euclidean embeddings \cite{DBLP:journals/corr/ChamberlainCD17}.

Given the success of hyperbolic embedding in providing faithful representations, it seems relevant to set up an approach that applies pure hyperbolic techniques in order to learn nodes and more generally communities representations on graphs. This motivation is in line with several recent works that have demonstrated the effectiveness of hyperbolic tools for different applications in Brain computer interfaces \cite{DBLP:journals/tbe/BarachantBCJ12}, Computer vision \cite{DBLP:journals/tit/SaidBBM17} and Radar processing \cite{DBLP:journals/jstsp/ArnaudonBY13}. In particular \cite{DBLP:journals/tbe/BarachantBCJ12} used the distance to the Riemannian barycenter on the space of covariance matrices to classify brain computer signals. \cite{DBLP:journals/tit/SaidBBM17} introduced Expectation-Maximisation (EM) algorithms on the same space then applied it to classify patches of images. \cite{DBLP:journals/jstsp/ArnaudonBY13} used the Riemannian median on the Poincar\'e disc to detect outliers in Radar data. 

Motivated by the success of hyperbolic embeddings on the one hand and hyperbolic learning algorithms on the other hand, we propose in this paper a new approach to the problem of learning nodes and communities representations on graphs. Two different applications are targeted:

\noindent\textbf{Unsupervised learning on GSD.} We propose to learn communities based on Poincaré embeddings \cite{NIPS2017_7213} and the recent formalism of Riemannian EM algorithms  \cite{DBLP:journals/tit/SaidHBV18}. Our approach can be considered as a Riemannian counterpart of \textit{ComE} \cite{cavallari2017learning}. 

\noindent\textbf{Supervised learning on GSD.} 
We propose a supervised framework that uses community-aware embeddings of graphs in hyperbolic spaces. To evaluate this framework we rely on three different tools to retrieve communities:  distance to the Riemannian barycenter, Riemannian Gaussian Mixture Models (GMM) \cite{DBLP:journals/tit/SaidHBV18}, or Riemannian logistic regression \cite{ganea2018hyperbolic}.

For both unsupervised and supervised frameworks, we evaluate the proposed methods on real-data social networks. We also make comparisons  with the $\textit{ComE}$ approach  \cite{cavallari2017learning} and recent geometric methods \cite{Cho2018LargeMarginCI}.

This paper is organised as follows. Section \ref{section_ball} reviews the geometry of the Poincar\'e ball and introductory tools which will be used after: Riemannian barycenter and Riemannian Gaussian distributions. In Section 3, we review Poincar\'e embedding \cite{NIPS2017_7213} and present our approach to learn GSD. Finally Section \ref{section_experiments} provides experiments and discusses comparisons with state-of-the-art. 
\section{The Poincaré Ball model: Geometry and tools  \label{section_ball}}
In this section, we first review the geometry of the Poincaré ball model as a hyperbolic space, that is a Riemannian manifold with constant negative sectional curvature. In particular, we recall expressions of the exponential and logarithmic maps which will be used in the sequel. Then, we review the definitions and some properties related to Riemannian barycenter and Riemannian Gaussian distributions. Our approach presented in the next section strongly relies on algorithms $K$-means and EM developed from these two concepts.

\subsection{Geometry of the Poincaré ball}

The Poincar\'e ball model of $m$ dimensions $(\mathbb B^m,g^{\mathbb B^m})$ is the manifold $\mathbb B^m=\{x\in\mathbb R^m : ||x||<1\}$ equipped with the Riemannian metric:

$$g^{\mathbb B^m}_x=\frac{4}{(1-||x||^2)^2} g^{\mathbb E}$$
where $g^{\mathbb E}$ is the Euclidean scalar product. The Riemannian distance induced by this metric is given by:
$$d(x,y)=\text{arcosh}\left( 1 + 2 \frac{||x-y||^2}{(1-||x||^2)(1-||y||^2)}\right)$$ 
Endowed with this metric, $\mathbb{B}^m$ becomes a hyperbolic space 
\cite{helgason,alekseevskij1993geometry}. \cite{ganea2018hyperbolic} gave explicit expressions for the Riemannian exponential and logarithmic maps associated to this distance as follows. First, define Möbius addition $\oplus$ for $x,y \in \mathbb{B}^m$ as:

$$x\oplus y=\frac{(1+2\langle x,y\rangle + ||y||^2)x+(1-||x||^2)y}{1+2\langle x,y\rangle + ||x||^2||y||^2}$$

For $x\in \mathbb B^m$ and $y\in\mathbb R^m\setminus\{0\}$, the exponential map is defined as:
$$\text{Exp}_x(y)=x\oplus \left(\text{tanh}\left(\frac{||y||}{1-||x||^2}\right) \frac{y}{||y||}\right)\  $$

\begin{figure}
\centering
    \includegraphics[scale=0.4]{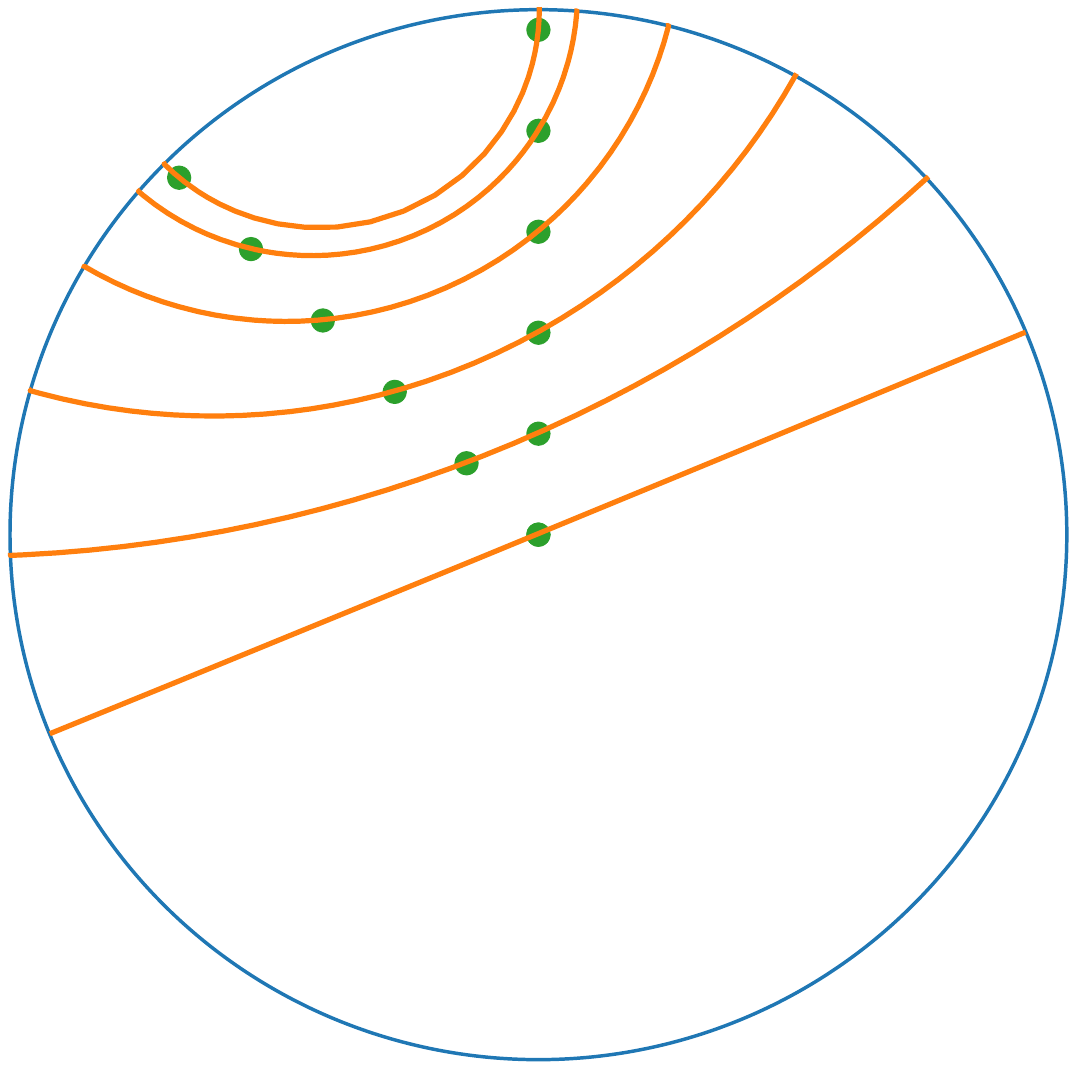}
    \caption{Geodesics passing through two points in the \textit{Poincaré disc}}
    \label{fig:geodesics_poincare}
\end{figure}

$\text{Exp}_x$ is a diffeomorphism from $\mathbb R^m$ to $\mathbb B^m$. Its inverse, called the logarithmic map is defined for all $x, y \in \mathbb{B}^m$, $ x\neq y$ by:

$$\text{Log}_x(y)=(1-||x||^2)\text{tanh}^{-1}\left(||-x\oplus y||\right)\frac{-x\oplus y}{||-x\oplus y||}\ $$ 
Figure \ref{fig:geodesics_poincare} shows different geodesics linking two points on $\mathbb{B}^2$.
Notice that geodesics passing through the center are straight lines and as points get closer to the boundary geodesics become more and more curved. 

\subsection{Riemannian barycenter}
As a Riemannian manifold of negative curvature, $\mathbb{B}^m$ enjoys the property of existence and uniqueness of the Riemannian barycenter \cite{afsarii}. More precisely, for every set of points $\{x_i, 1\le i\le n\}$ in $\mathbb B^m$, the empirical Riemmanian barycenter 
$$\hat{\mu}_n=\text{argmin}_{\mu \in \mathbb B^m} \left(\sum_{i=1}^n d^2(\mu,x_i)\right)$$
exists and is unique. Several stochastic gradient algorithms can be applied to numerically approximate $\hat{\mu}_n$ \cite{bonnabel,arnaudonstoch,arnaudon1,DBLP:journals/jstsp/ArnaudonBY13,miclo}. Riemannian barycenter has given rise to several developments in particular, $K$-means and EM algorithms which will be used in this paper.

\subsection{Riemannian Gaussian distributions}\label{sec}

Gaussian distributions have been extended to manifolds in various ways \cite{skovgaard1984riemannian,pennec2006intrinsic,DBLP:journals/tit/SaidHBV18}. In this paper we rely on the recent definition provided in \cite{DBLP:journals/tit/SaidHBV18} 
as it comes up with an efficient learning scheme based on Riemannian mixture models. 
In the next section, these models will be applied to learn  communities on graphs. The same distribution was particularly used in  \cite{DBLP:journals/corr/abs-1901-06033} to generalise variational-auto-encoders to the Poincaré Ball. Moreover \cite{ovinnikov2019poincar} proposed an extension of Wasserstein auto-encoders to manifolds based on the same definition of Gaussian distributions.

Given two parameters $\mu\in \mathbb B^m$ and $\sigma>0$, respectively interpreted as theoretical mean (or barycenter) and standard deviation, the Riemannian Gaussian distribution $\mathcal G(\mu,\sigma)$ on $\mathbb B^m$, is given by its density: 
$$f(x|\mu,\sigma)=\frac{1}{\zeta_m(\mu,\sigma)} \exp\left[-\frac{d^2(x,\mu)}{2\sigma^2}\right]$$
with respect to the Riemannian volume $dv(x)$ \cite{ DBLP:journals/tit/SaidHBV18}. A first interesting and practical property is that $\zeta_m(\mu,\sigma)$ does not depend on $\mu$:

$$\zeta_m(\mu,\sigma)=\zeta_m(0,\sigma)=\zeta_m(\sigma)=\int_{\mathbb B^m} \exp\left[-\frac{d^2(x,0)}{2\sigma^2}\right] dv(x)$$

\begin{figure}
    \centering
    \includegraphics[scale=0.4]{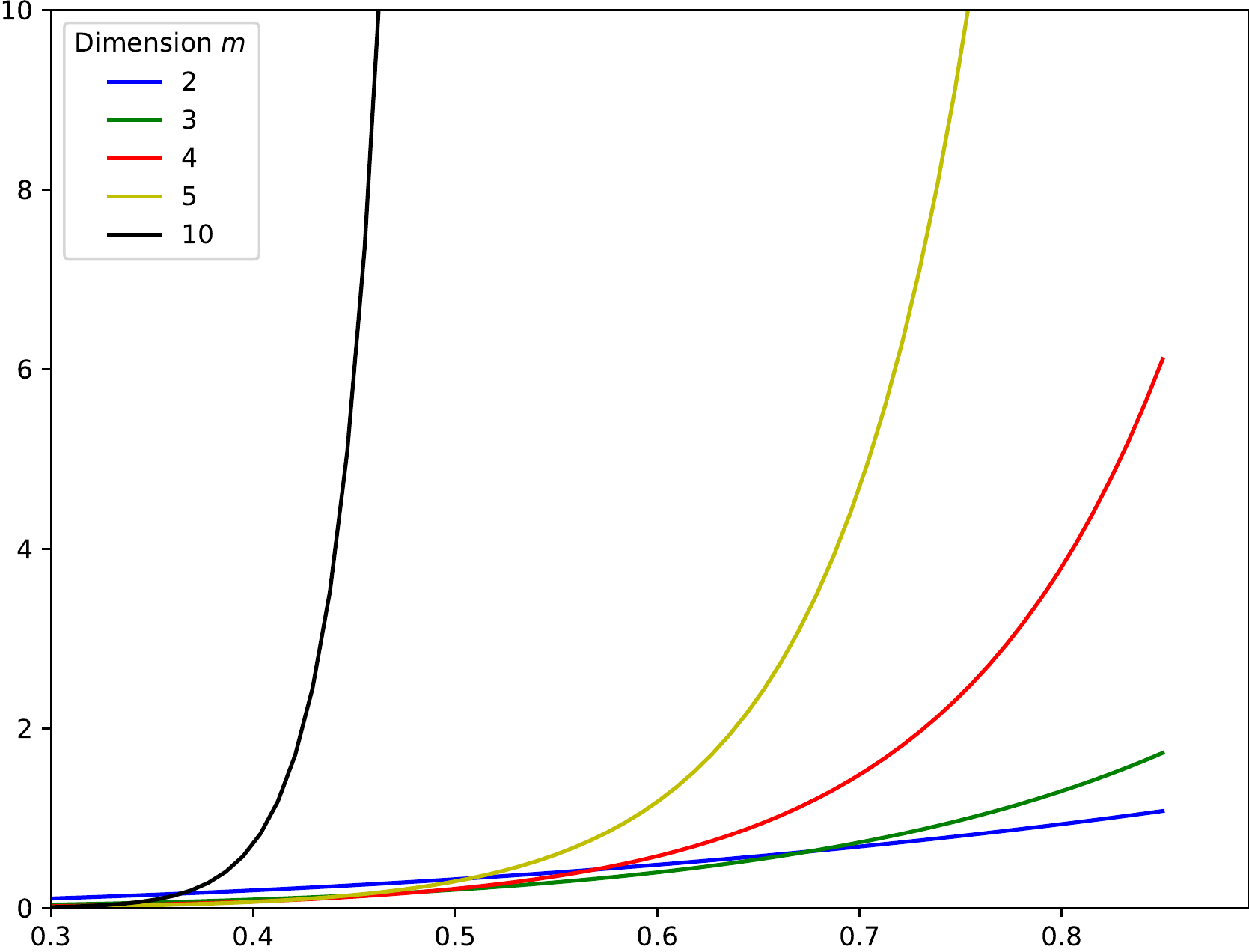}
    \caption{Plots representing $\sigma\mapsto \zeta_m(\sigma)$ for some values of the dimension $m$ of the Poincaré ball}
    \label{fig:zeta}
\end{figure}

A more explicit expression of $\zeta_m(\sigma)$ has been given recently in \cite{DBLP:journals/corr/abs-1901-06033} as follows:
$$     \zeta_m(\sigma) = \sqrt{\frac{\pi}{2}}\frac{\sigma}{2^{m-1}} \sum\limits^{m-1}_{k=0} (-1)^k C^{k}_{m-1} e^{\frac{p_k^2\sigma^2}{2}}\bigg( 1+\text{erf}(\frac{ p_k\sigma}{\sqrt{2}})\bigg)$$

with $p_k =(m-1) - 2k$. Figure \ref{fig:zeta} displays some plots of the function $\sigma\mapsto \zeta_m(\sigma)$. 

Recall Maximum Likelihood Estimation (MLE) of the parameters $\mu$ and $\sigma$, based on independent samples $x_1, \ldots, x_n$ from $\mathcal G(\mu,\sigma)$ given as follows: 

\begin{itemize}
    \item The MLE $\hat{\mu}_n$ of $\mu$ is the Riemannian barycenter of $x_1, \ldots, x_n$. 
    \item The MLE $\hat{\sigma}_n$ of $\sigma$ is 
    $$\hat{\sigma}_n=
    \Phi\left( \frac{1}{n} \sum^{\scriptscriptstyle n}_{i=1} d^2(\hat{\mu}_n,x_i) \right)$$
    where $ \Phi: \mathbb R_+\longrightarrow \mathbb R_+ $ is a strictly increasing bijective function given by the inverse of $\sigma\mapsto \sigma^3 \times \frac{d  }{\mathstrut d\sigma}\log \zeta_m(\sigma)$.
\end{itemize}

\section{Learning hyperbolic community embeddings\label{section_approach}}

This section generalises the mechanism of \textit{ComE} to the Riemannian setting. Our approach is based on hyperbolic embedding and Riemannian mixture models. 

Consider a graph $G(V,E)$ where $V$ is the set of nodes and $E \subset V\times V$ the set of edges. The goal of embedding GSD is to provide a faithful and exploitable representation of the graph structure. It is mainly achieved by preserving  \textit{first-order} proximity that enforces nodes sharing edges to be close to each other. It can additionally preserve \textit{second-order} proximity that enforces two nodes sharing the same context (i.e., nodes that are neighbours but not necessarily directly connected) to be close.

\paragraph{First-order proximity.}
To preserve first-order proximity we adopt a loss function similar to \cite{NIPS2017_7213}:
$$
    O_1 = -\sum \limits_{(v_i, v_j) \in E} log(\sigma(-d^2(\phi_i, \phi_j)))
$$

with $\sigma(x)=\frac{1}{1+e^{-x}}$ is the sigmoid function and $\phi_i \in \mathbb{B}^m$ is the embedding of the $i$-th node of $V$.

\paragraph{Second-order proximity.}
In order to preserve second-order proximity,
the representation of a node has to be close to the representations of its context nodes.  For this, we adopt the negative sampling approach \cite{NIPS2013_5021} and consider the loss:
$$     O_2 = - \sum\limits_{v_i\in V} \sum\limits_{v_j \in C_i} \bigg[ log(\sigma(-d^2(\phi_i, \phi_j'))) + \sum \limits_{v_k\sim \mathcal{P}_n} log(\sigma(d^2(\phi_i, \phi_k')))  \bigg]$$

with $C_i$ the nodes in the context of the $i$-th node, $\phi_j'\in \mathbb{B}^m$ the embedding of $v_j\in C_i$ and $\mathcal{P}_n$ the negative sampling distribution over $V$: $\mathcal{P}_n(v)=\frac{deg(v)^{3/4}}{\sum\limits_{v_i\in V}deg(v_i)^{3/4}}$. Following the \textit{ComE} approach, we introduce in the next paragraph a third objective function in order to improve the mechanism of community detection and embedding.

\subsection{Expectation-Maximisation for hyperbolic GMM \label{EM}}
Riemannian EM algorithm introduced in \cite{DBLP:journals/tit/SaidHBV18} has similarities with the usual EM on Euclidean spaces. It is employed to approximate distributions on manifolds by a mixture of standard distributions. We recall the density $F$ of a GMM on the Poincaré ball 
$$F(x|\mu,\sigma)= \sum_{k=0}^K\pi_k f(x|\mu_k,\sigma_k)$$
where $x\in \mathbb{B}^m$, $\pi_k$ the mixture coefficient, $\mu_k$, $\sigma_k$ the parameters of the $k$-th Riemannian Gaussian distribution.
The Riemannian EM algorithm computes the GMM that best fits a set of given points $x_1,\cdots,x_N$. This is done by maximising the log-likelihood of the joint probability under the hypothesis that observations are independent. Given an initialisation of $\pi$, $\mu$ and $\sigma$, performing EM numerically translates to alternating between Expectation and Maximisation steps. Fitting a Gaussian mixture model on Riemannian manifolds have some similarities with its Euclidean counterpart. The main complexity is that there are no known closed forms for the log-likelihood maximisation of the mean and standard deviation.

\subsubsection{Expectation}
The expectation step estimates the probability of each sample $x_i$ to belong to a particular Gaussian cluster. The estimation is based on the posterior distribution $\mathbb{P}(z_i=k|x_i)$, i.e. the probability that the sample $x_i$ is drawn from the $k$-th Gaussian distribution: 

$$w_{ik}:= \mathbb P(z_i=k| x_i) = \frac{\pi_k \times f(x_i| \mu_k, \sigma_k)}{\sum \limits_{j=0}^N \pi_k \times f(x_j| \mu_k, \sigma_k)}$$

In this expression $z_1,\cdots,z_N$ represent the latent variables associated to the mixture model. 

\subsubsection{Maximisation \label{maximisation}}\label{max}
The maximisation step estimates the mixture coefficient $\pi_k$, mean $\mu_k$ and standard deviation $\sigma_k$ for each Gaussian component of the mixture model.

\paragraph{Mixture coefficient.}
The mixture coefficients are computed by :
$$ \pi_k = \frac{1}{N} \sum \limits_{i=1}^N w_{ik}$$

\paragraph{Mean.} Updating the $\mu$ parameter relies on estimating weighted barycenters. For this, it is required to approximate the weighted barycenter $\hat{\mu_k}$ of the $k$-th cluster:
$$\hat{\mu_k} = \arg\min \limits_\mu \sum\limits_{i=1}^N w_{ik}d^2(\mu, x_i)$$
via Riemannian optimisation. We will use Algorithm \ref{alg:barycenter} of \cite{DBLP:journals/jstsp/ArnaudonBY13} which has proven effective for Radar applications. This algorithm returns an estimate of $\hat{\mu}_k$. 

\begin{algorithm}[!ht]    

    \caption{Barycenter computation \label{gsd_barycenter}}
    \label{alg:barycenter}
    \textbf{Require : }
                        $W=(w_{ik})$ weight matrix,  
                        $\{x_1,\cdots,x_N\}$ a subset of $\mathbb{B}^m$,
                        $\epsilon$ convergence rate (small),
                        $\lambda$ barycenter learning rate
    \begin{algorithmic}[1]
    \State Initialisation of $\mu_k^{0}$
        \Do
            \State $\mu_k^{t+1} \gets \text{Exp}_{\mu_k^{t}}\bigg(\lambda  \frac{2}{\sum\limits_{i=1}^N wik}\sum\limits_{i=1}^N w_{ik} \text{Log}_{\mu_k^t}(x_i)\bigg)$   
       \doWhile{$d(\mu_k^{t}, \mu_k^{t+1}) > \epsilon$}
\Comment{$t$ is the iteration index}
    \end{algorithmic}
    \Return $\mu_k^{t+1}$
\end{algorithm}

\paragraph{Standard deviation.} Recall the MLE of the standard deviation of the Gaussian distribution is previously defined in Section \ref{sec}. Thanks to the property of $\Phi$ being strictly increasing and bijective, estimation of $\sigma_k$ can be given by solving the following problem :
$$\hat{\sigma_k}=\argmin\limits_{\sigma_s} \left| \left( \frac{1}{\sum\limits_{i=0}^N w_{ik}} \sum\limits_{i=0}^N d^2(\mu_k, x_i)w_{ik}\right) - \Phi^{-1}(\sigma_s) \right|$$

A grid-search is used to find an approximation of $\sigma_k$ by computing its values for a finite number of $\sigma_s$. To compute $\Phi^{-1}(\sigma_s)$ which involves the term $\frac{d}{d \sigma}log\zeta_m(\sigma)$ we use automatic differentiation algorithm provided by the back-end.

\subsection{Learning with communities}

\label{loop}

Since we want to learn embeddings for the purpose of community detection and classification, we need to ensure that embedded nodes belonging to the same community are close. Therefore, similarly to the statement made in \textit{ComE} \cite{cavallari2017learning} we connect node embedding (first and second-order proximity via the minimisation of $O_1$ and $O_2$) together with community awareness using a loss function $O_3$. The later is named the community loss, which we write as:
$$
    O_3 = - \sum \limits_{i=1}^{|V|}\sum \limits_{k=0}^K w_{ik}log\left( \frac{1}{\zeta_m(\sigma_k)}e^{-\frac{d^2(x_i, \mu_k)}{2\sigma_k^2}} \right)
$$
To solve community and graph embedding jointly we therefore minimise the final loss function:

$$L = \alpha.O_1 + \beta. O_2 + \gamma.O_3$$

with $\alpha, \beta, \gamma$ the weights of respectively first, second-order and community losses. The optimisation is done iteratively as follows:  
\begin{enumerate}
    \item Update embeddings for all edges based on $O_1$.
    \item Generate random walks using   \textit{DeepWalk} \cite{Perozzi:2014:DOL:2623330.2623732} and then update embeddings for selected nodes based on $O_2$. 
    \item Update embeddings to fit the GMM based on $O_3$.
    \item Update the GMM parameters $\mu$, $\sigma$ and $\pi$ as described in Section \ref{EM}.
\end{enumerate}

\paragraph{Riemannian Gradient Descent (RGD).}
Following the idea of \cite{ganea2018hyperbolic}, we use the RGD Algorithm, which first computes the gradient on the tangent space and then project the new values on the ball
$$\phi^{t+1} =  \text{Exp}_{\phi^t}\left(\lambda \frac{\partial h}{ \partial \phi^t}\right)$$

with $h$ the function to optimise and $\lambda$ a learning rate.

\subsection{Complexity and Scalability}

\paragraph{Complexity. } With $|V|$ the number of nodes and $|E|$ the number of edges, the time complexity of embedding GSD for first and second-order proximity are respectively in $\mathcal{O}(|E|)$ and $\mathcal{O}(|V|)$. Assuming a fixed number of iterations for all RGD computations, performing embedding and Riemannian barycenter based $K$-means has $\mathcal{O}(|V|.m.K +|E|)$ complexity thus is in $\mathcal{O}(|V|+|E|)$ for large graphs. Similarly, the time complexity of community embedding with EM loop is in $\mathcal{O}(|V|+|E|)$.

\paragraph{Scalability. }
\label{para:scalability}
The GSD embedding update process for $O_1$ and $O_2$ in $\mathbb B^m$ is a linear function of $m$. Therefore, embedding the GSD scales well to large datasets.
To accelerate updates of $O_1, O_2, O_3$, we use batch gradient descent algorithm mainly for large datasets.  

Although run-times of Riemannian $K$-means and EM are higher than their Euclidean versions (mainly due to RGD), scalability is not affected by the change of the underlying space: each operation has higher run-time but the total number of operations as function of the dataset size scales similarly to Euclidean spaces. Notice that the computation of the normalisation factor $\zeta_m (\sigma)$ makes fitting hyperbolic GMM computationally challenging, this can be visually perceived in Figure \ref{fig:zeta}. Thus, numerically we have to deal with the out of bound numbers due to the terms $e^{\frac{(m-1)^2\sigma^2}{2}}$. To keep the computation time-bounded, we fixed a finite number of values for $\sigma$ for which we pre-computed $\zeta_m$. To avoid the out of bound issue we restricted our work to $10$ dimensions while increasing the floating-point precision (see Section \ref{section_experiments}).
\section{Experiments \label{section_experiments}}

In this section, we provide experimental results and compare them with recent works from the literature\footnote{The package implementing our algorithms will be made publicly available online in the near future.}.
To assess the relevance of the learned embeddings we designed an unsupervised community detection and a supervised node classification experimental frameworks.
 We present experiments on \textit{DBLP}\footnote{https://aminer.org/billboard/aminernetwork} a library of scientific papers graph, \textit{Wikipedia}\footnote{http://snap.stanford.edu/node2vec/POS.mat} a graph built on Wikipedia dump, \textit{BlogCatalog}\footnote{http://socialcomputing.asu.edu/datasets/BlogCatalog3} a blog social network graph and \textit{Flickr}\footnote{http://socialcomputing.asu.edu/datasets/Flickr} a graph based on Flickr users.
We also experiment on several low scale datasets shown in Table \ref{tab:datasets} and provide visualisations of the learned embeddings when the dataset allows it.

\begin{table}[h]
    \caption{Characteristics of the datasets used for experimental evaluation. $|V|$ the number of nodes, $|E|$ the number of edges, $K$ the number of communities and ML whether or not the dataset is multi-label (whether or not a node can belong to several communities).\\\label{tab:datasets}}
\centering

    \begin{tabular}{ l|c|c|c|c}
         \hline
         Corpus  & $|V|$ & $|E|$ & $K$ & ML\\
         \hline
         \hline
     Karate & 34   & 77 &2 & no\\ 
     Polblogs& 1224 & 16781&2 & no\\
     Books& 105  & 441&3&no\\ 
     Football &  115 & 613 &12&no\\ 
    \hline
        DBLP& 13,184 & 48,018 & 5&no\\
        Wikipedia & 4,777 & 184,812 & 40&yes\\
        BlogCatalog& 10,312 & 333,983 & 39&yes\\
        Flickr&80,513&5,899,882&195&yes\\
         \hline
    \end{tabular}

    \label{corpus}
\end{table}
\paragraph{Hyper-parameters selection.}
We selected hyper-parameters that give the best cross-validation performances in terms of conductance, defined below, when running our algorithm and the baseline \textit{ComE}. In particular: $\lambda$ the learning rate, $c$ the size of the context window and $t$ the number of negative sampling nodes which appeared to be  the most influencing on results. For our algorithm
\footnote{For Flickr dataset we only run one experiment for each parameter, moreover, the number of parameters tested is lower than those tested for others datasets}, parameters $c$ and $t$ were selected from the set $\{5, 10\}$, $\beta$ and $\alpha$ $ \in \{0.1, 1\}$, $\gamma \in \{0.01, 0.1\}$ and $\lambda$ from $[1e-2,1e-4]$. For all experiments, we generated, for each node, $10$ random walks, each of length $80$. For the first 10 epochs, embeddings are trained using only $O_1$ and $O_2$ and then using the complete loop as described in Section \ref{loop}.

As for running the algorithm \textit{ComE}, the tested parameters are $\lambda \in \{0.1, 0.01\}$, $\beta \in \{1, 0.1\}$ and $\gamma \in \{0.1, 0.01\}$. All others parameters are the ones set by default.\footnote{\textit{ComE} code available at https://github.com/vwz/ComE}

\subsection{Unsupervised Clustering and Community Detection}
In this section we present the unsupervised experiments based on the EM algorithm.

\paragraph{Unsupervised Community Detection.}

\begin{table*}[t]
    \footnotesize
    \caption{Unsupervised community detection performances for Hyperbolic Gaussian Mixture Model (H-GMM) in comparison with state-of-the-art method \textit{ComE}.}
    \vspace{3pt}
    \centering
    \footnotesize
    \begin{tabular}{lc|cc|cc|cc}
        \hline

        & & \multicolumn{2}{c|}{Precision@1}&\multicolumn{2}{c|}{Conductance } &\multicolumn{2}{c}{NMI}  \\
        \hline
        \hline
        Dataset & D & H-GMM & ComE &  H-GMM & ComE & H-GMM & ComE \\
        \hline
        
        \multirow{3}{*}{DBLP} & 2 &$\mathbf{79.3} \pm 3.3$ & $75.9$ & $\mathbf{6.9}\pm1.2$ & 10.1 & $\mathbf{57.6}\pm2.5$ & $55.9$\\
           & 5 & $82.0\pm3.2$ & $\mathbf{87.5}$ & $\mathbf{5.4}\pm0.2$ & $5.8$& $58.0\pm1.2$ & $\mathbf{62.0}$\\
           & 10 & $\mathbf{81.4}\pm1.9$ & $80.4$   & $\mathbf{5.5}\pm0.2$ & $\mathbf{5.6}$& $\mathbf{57.8}\pm 0.3$ & $56.6$\\
        \hline

        \multirow{3}{*}{Wikipedia}  & 2 &5.8$\pm$1.1 &
        5.8& 97.1$\pm$3.0 & \textbf{94.8}& 6.5$\pm$1.4& \textbf{6.8}\\
           & 5 & 5.1$\pm$1.2& \textbf{6.7} &94.8$\pm$4.3 & \textbf{90.7} & \textbf{8.6}$\pm$0.5& 8.4\\
           & 10 & 5.6$\pm$1.3 & \textbf{6.3} &92.0$\pm$4.9& \textbf{89.4} & \textbf{8.5}$\pm$0.5 & 8.4\\
        \hline
        \multirow{3}{*}{BlogCatalog} & 2 & \textbf{6.3}$\pm$0.4 & 4.3
           & \textbf{94.7}$\pm$5.1&  94.8 & 2.8$\pm$0.1 & \textbf{3.3}
           \\
           & 5 &5.5$\pm$0.7& \textbf{7.3} & 91.9$\pm$6.7  & \textbf{89.0}& 8.0$\pm$0.6 & \textbf{11.0}\\
           & 10 & \textbf{7.3}$\pm$0.9 & 7.0 & \textbf{86.4}$\pm$6.9 & 87.9 & \textbf{13.6}$\pm$0.3& \textbf{13.7} \\
        \hline
        \multirow{2}{*}{Flickr} 
        & 2  &  \textbf{9.4}$\pm$0.3 & 3.7 & \textbf{96.3}$\pm$11.1 & \textbf{96.5} & 21.4 &  \textbf{23.1}$\pm$0.3 \\
        & 5  &  \textbf{8.0}$\pm$0.4 & 5.5 & 92.6$\pm$12.2 & 91.6 & \textbf{29.9} &  \textbf{30.0}$\pm$0.0 \\
          & 10 & \textbf{9.0}$\pm$0.3 & 8.0 & \textbf{89.0}$\pm$13.1 & \textbf{89.3} & \textbf{33.0}$\pm$0.1 & \textbf{33.1}\\
        \hline
    \end{tabular}

    \label{tab:unsup}
    \vspace{-10pt}

\end{table*}

We perform optimisations of $O_1, O_2$ and $O_3$ for a fixed number of iterations. Then, the probability for a node to belong to each Gaussian cluster is computed. Finally, each node is labelled with the community giving it the highest posterior probability. The results are shown in Table \ref{tab:unsup} and the evaluation metrics are detailed next. 

\paragraph{Unsupervised evaluation metrics.}
For both experiments we evaluate the results using \textit{Conductance}, Normalised Mutual Information \textit{NMI}, and \textit{Precision@1}.\\

$\bullet$ \textit{Conductance:} measures the number of edges shared between separate clusters. The lower the conductance is, the less edges are shared between clusters. 
    Let $C_i$ be the set of nodes for the cluster $i$, $A$ the adjacency matrix of the GSD, the mean conductance $MC$ over clusters is given by:
        $$
            MC = \frac{1}{K}\sum \limits_{i=1}^{K} \frac{\sum\limits_{j\in C_i, k \notin C_i} A_{jk}}{\min \left(\sum\limits_{j\in C_i, k\in V }A_{jk},\sum\limits_{j\notin C_i, k\in V }A_{jk} \right)}
        $$

$\bullet$ \textit{NMI : } Let $A_{ij}$ be the number of common nodes belonging to both the predicted cluster $i$ and the real cluster $j$, $|V|$ being the number of nodes, $A^p_i$ the number of elements in the $i$-th predicted cluster and $A^t_i$ the number of elements in the $i$-th real community. The NMI is given by: 
    $$
    \text{NMI} = \frac{-2\sum\limits_{i=0}^K\sum\limits_{j=0}^K  A_{ij} log\left(\frac{A_{ij}|V|}{A^p_iA^t_j}\right)}
    {\sum\limits_{i=0}^K A^p_i log \left(\frac{A^p_i}{|V|}\right) + \sum\limits_{j=0}^K A^t_j log\left(\frac{A^t_j}{|V|}\right)}
    $$
$\bullet$ \textit{Precision@1}:
Since the real labels of  communities are known, we propose a supervised measure derived from the precision metric that uses the true community labels.
A problem we encountered is that the associations between predicted labels and the true ones are unknown. For small number of communities, all possible combinations are computed and the best performance is reported.
This solution is not tractable when the number of communities grows. A greedy approach is then used: the predicted labels of the largest cluster is first associated to the known true labels of the dominant class, similarly the second largest cluster is associated to the second dominant class with known labels and so on. The process is repeated until all clusters are treated. Finally, for mono-label datasets, the \textit{Precision@1} corresponds to the mean percentage of the correctly guessed labels. For  multi-label datasets, an element is considered correctly labelled if the inferred community corresponds to one of its true known communities.

\subsection{Supervised Classification}

\begin{table*}[t]
    \footnotesize
    \centering
    \caption{Results for the different classification methods. With : H-$K$-Means the supervised hyperbolic $K$-Means; H-GMM the supervised hyperbolic Gaussian mixture model; H-LR the regression logistic in hyperbolic space; \textit{ComE} \cite{cavallari2017learning}.}
    \label{table:supervised}
    \begin{tabular}{c|c|c|c|c|c}
       \hline
       Dataset & Dim & H-$K$-Means & H-GMM & H-LR & \textit{ComE} \\
       \hline
       \hline
       \multirow{4}{*}{DBLP}& 2 &85.1$\pm$1.8&\textbf{87.4}$\pm$1.5 &79.3$\pm$11.& 75.3\\
       & 5 &89.6$\pm$0.7&90.5$\pm$0.3&\textbf{91.1}$\pm$0.5&89.8\\
       & 10 &90.0$\pm$0.5&90.6$\pm$0.5&\textbf{91.6}$\pm$0.6&90.1\\
       \hline
       \multirow{4}{*}{Wikipedia}& 2 & 0.4/0.2/0.1& 44.1/26.3/19.1 & 46.8/28.0/20.7&\textbf{47.2/28.5/21.1} \\
       & 5 & 3.4/1.2/0.7 & 42.9/26.8/19.9 & \textbf{47.1/29.0/21.6}&\textbf{47.2/28.7/12.8}\\
       & 10 & 7.0/2.3/1.4 & 43.4/27.4/20.1& \textbf{48.0/29.7/21.9} &\textbf{48.1/29.6/22.1} \\

       \hline
       \multirow{4}{*}{BlogCatalog}& 2 & 3.2/2.4/3.8& 16.2/11.9/10.4& \textbf{16.2/12.3/10.7}& \textbf{16.2/12.5/10.7}\\
       & 5 & 9.8/4.7/4.8 & 22.1/14.1/11.4& 23.0/14.9/12.0& \textbf{25.9/16.4/12.8}\\
       & 10 & 21.3/8.5/6.8 & 32.3/18.8/14.1& \textbf{34.3/19.6/14.6}&  34.1/19.3/14.4\\
      \hline
      \multirow{2}{*}{Flickr}
      & 2 & 4.1/1.9/1.4 &\textbf{22.5/14.0/10.5}& 20.3/11.5/8.7 & 20.3/12.3/9.3\\
      & 5 & 9.9/3.9/2.5 &25.4/15.8/12.1& \textbf{26.8/16.0/12.1}& 26.6/15.5/11.7\\
      & 10 & 13.8/5.1/3.3&25.5/15.9/12.3 & \textbf{32.2/18.5/13.7}& 31.1/17.9/13.3\\
    \hline

    \end{tabular}

\end{table*}

In this section, experiments in the supervised framework are presented. Assuming to know the communities of some nodes, the objective is to predict the communities of unlabelled nodes. Three different methods (detailed in the next paragraphs) are used to predict labels: Riemannian $K$-Means, GMM or Logistic Regression (LR) based on geodesics. For all experiments we apply a $5$-cross-validation process with $20\%$ of the dataset for validation. The \textit{Precision@1} is reported for mono-label datasets (i.e., each element belongs to a unique community), additionally \textit{Precision@3} and \textit{5} are reported for multi-label datasets (i.e., each element can belong to several communities). Table \ref{table:supervised} reports the mean and standard deviation of the  $\textit{Precision}$ over 5 runs.

\paragraph{Supervised $K$-Means.}
This method uses a  supervised version of $K$-Means by computing the barycenter of nodes belonging to a given community. Each barycenter is considered as a cluster centroid representing the community. Then nodes from the validation set are associated to the communities of the $n$ nearest barycenters (depending on the considered \textit{Precision@$n$}).

\paragraph{Supervised GMM.} Given the known labels of the train data, a GMM is estimated. 
 The parameters of the GMM are obtained by applying one maximisation step given the train nodes (Section \ref{maximisation}) using the following estimate:  $$w_{ik} = \frac{{y_{i,k}}}{\sum\limits_{k=0}^Ky_{i,k}}$$
 with $y_{i,k}=1$ if node $i$ belongs to the community $k$ and $y_{i,k}=0$ otherwise.
 To predict the community $\hat{k}$ of a given node embedded as $x_i \in \mathbb{B}^m$, we use Bayes decision rule: 
 $$\hat{k}=\text{argmax}_k\mathbb P(k|x_i)= \text{argmax}_k w_{ik} f(x_i|\hat{\mu_k},\hat{\sigma_k})$$
 with $\hat{\mu_k}, \hat{\sigma_k}$ the estimated parameters of the $k$-th Gaussian distribution.



\paragraph{Logistic Regression.}
In order to further compare \textit{ComE} and its Riemannian version developed in this paper, we propose similarly to \cite{cavallari2017learning} to learn a classifier. For the baseline \textit{ComE}, we use the usual Euclidean logistic regression. For its Riemannian version we rely on the hyperbolic logistic regression proposed in \cite{ganea2018hyperbolic}. From this paper, recall that for a given hyper-plane $H_{a,p}$ defined by a point $p \in \mathbb{B}^m$ and a tangent vector $a \in T_p \mathbb{B}^m$, 
the distance of a given 
$x \in \mathbb{B}^m$ to
$H_{a,p}$ is:
$$
d(x, H_{a,p})= sinh^{-1}\left( \frac{2|<-p\oplus x, a_k>|}{(1-||-p\oplus x||^2)||a_k||}\right)
$$
Then, the probability of a node $x$ to belong to the community $k$ is modelled by 
$$
\mathbb{P}(z=k| x) = \sigma \big( sign(-p\oplus x)d(x, H_{a,p})\big)
$$ with $z$ the community latent variable.

\begin{table}[!h]
    \footnotesize
    \centering
    \caption{Performances obtained by our method compared to Hyperbolic-SVM\cite{Cho2018LargeMarginCI} (H-SVM) for small scale datasets for 2-dimensional embedding. Results are the means for $5$-folds cross-validation for $5$ experiments }
    \begin{tabular}{c|c|c|c|c}
        \hline
         Dataset &H-SVM&H-$K$-Means&H-GMM&H-LR\\
         \hline
         \hline
         Karate& 86$\pm$3&\textbf{92}$\pm$11.&91$\pm$11.&87$\pm$15.\\
         Polblogs& 93$\pm$1&95$\pm$0.9&95$\pm$0.9&\textbf{96}$\pm$1.2\\
         Books & 73$\pm$4&83$\pm$8.3&\textbf{83}$\pm$7.5&82$\pm$7.4\\
         Football& 24$\pm$3&\textbf{80}$\pm$7.8&79$\pm$8.0&39$\pm$11.\\
    \end{tabular}

    \label{tab:small_dataset}
\end{table}

\paragraph{Hyperbolic SVM Comparison.}
To asess the relevance of the proposed classifier approaches, we compare in Table \ref{tab:small_dataset} our results with \textit{Hyperbolic} SVM \cite{Cho2018LargeMarginCI}. The performances are reported for embeddings of small datasets in two dimensions, using $5$ folds cross-validation. In terms of hyper-parameters, the context size is set to $3$ for \textit{Football} and $2$ for the other datasets. The reported results are directly taken from the H-SVM paper (where $5$ folds cross-validation are used as well).

\begin{figure}[!h]
    \centering
    \begin{subfigure}{.33\textwidth}
      \centering
            \includegraphics[scale = 0.5]{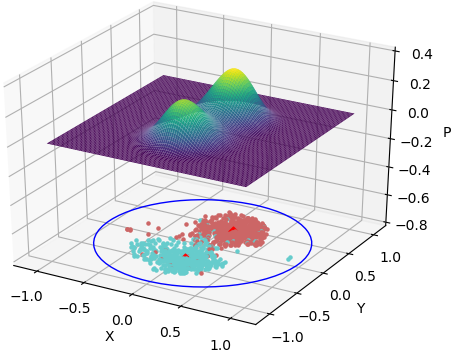}
      \label{fig:sub-first}
      \caption{Polblogs}
    \end{subfigure}
    \begin{subfigure}{.33\textwidth}
      \centering
            \includegraphics[scale = 0.5]{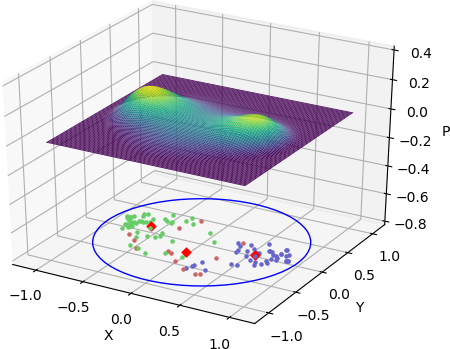}
      \label{fig:sub-first}
      \caption{Books}
    \end{subfigure}
    \begin{subfigure}{.33\textwidth}
      \centering
            \includegraphics[scale = 0.5]{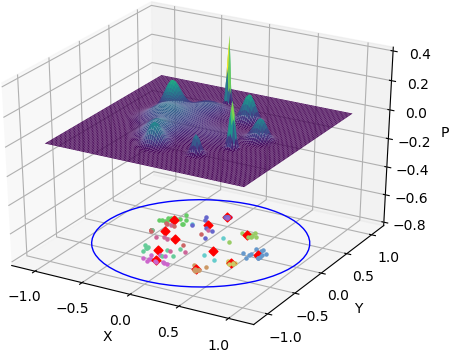}
      \label{fig:sub-first}
      \caption{Football}
    \end{subfigure}
    \caption{Visualization of Polblogs, Books and Football embeddings with their associated GMM}
    \label{fig:em_visu}  
\end{figure}

\subsection{Results and discussion}


\paragraph{Unsupervised community detection results.} 
Table \ref{tab:unsup} shows the performances of unsupervised community detection experiments. 
For \textit{DBLP}, results are better for all evaluation metrics in 2 dimensions compared to $\textit{ComE}$ and for precision and NMI in 10 dimensions. Furthermore, the conductances for our model in 10 dimensions and $\textit{ComE}$ in 128 dimensions are similar for this particular dataset. Our results exceed those of the baseline for \textit{Flickr} in terms of precision for all dimensions.
For the other datasets, performances of the two approaches are comparable. This shows that despite the constraint of using low dimensions, our approach is as competitive as the state-of-the-art approach \label{para:scalability}.

H-$K$-Means and H-GMM outperform H-SVM and H-LR when the number of communities is large (e.g., the \textit{football} dataset). We empirically show that geodesics separators are not always suited for classification problems. Differences in results between H-SVM and H-LR are mainly due to the quality of the learned embeddings. Indeed, \cite{Cho2018LargeMarginCI} uses embeddings preserving first and second-order proximity only.  

Figure \ref{fig:em_visu} presents visualisation of the learned embeddings with the known true labels of nodes (in colors), barycenters  of each community (red squares) and the associated GMM components.


\paragraph{Supervised community detection results.}
Table \ref{table:supervised} shows that the proposed embedding method obtains better or similar performances for the different datasets in low dimensions than \textit{ComE}. 
\textit{GMM} is particularly advantageous for \textit{DBLP} and \textit{Flickr} with superior performances in only two dimensions.
Using Riemannian logistic regression outperforms the baseline for \textit{Flickr} and \textit{DBLP} in 5 dimensions and reaches similar performances for the rest of the datasets, demonstrating the effectiveness of our method to learn community representations.  Notice that for \textit{DBLP}, results of \textit{ComE} in $128$ ($~92\%$) dimensions are comparable to our results in only $10$ dimensions. However, for Wikipedia and BlogCatalog, \textit{ComE} remains better in $128$ dimensions reaching respectively $49\%$ and $43\%$ of \textit{Precision@1}.


\section{Conclusion}
In this paper, we presented a methodology that combines embeddings together with clustering techniques on hyperbolic spaces to learn GSD. To this end, we designed a framework that uses Riemannian versions of $K$-means and Expectation-Maximisation jointly with GSD embedding.
We proposed a Riemannian analogue of state-of-the-art community learning algorithm \textit{ComE}.
Several experiments on real, large and referenced datasets showed better or similar results with state-of-the-art (\textit{ComE} and Hyperbolic SVM). This demonstrates empirically the effectiveness of our approach and encourages its use for more applications. 


\clearpage
\bibliographystyle{apalike}
\bibliography{biblio}
\appendix

\onecolumn
\appendix

\section{Riemannian optimisation}
In this section, the gradient computations for the loss functions $O_1$, $O_2$ and $O_3$ presented in the paper are shown. This is done by applying RGD to update nodes and context embeddings as follows:
$$
  \phi^{t+1} = \text{Exp}_{\phi^t}\left(-\eta \frac{\partial O_{\{1,2,3\}}}{\partial \phi}\right)
$$
where $\phi$ is a parameter, $t\in\{1,2,\cdots\}$ is the iteration number and $\eta$ is a learning rate.

Recall from \cite{DBLP:journals/jstsp/ArnaudonBY13} the formula giving the gradient of $d^p(\phi_i, \phi_j)$ where $\phi_i,\phi_j \in \mathbb{B}^m$ : 
\begin{equation}
    \label{eq:distance_hyperbolic_gradient}
    \nabla _{\phi_i} d^p(\phi_i,\phi_j)=-p* d^{p-1}(\phi_i,\phi_j)*\frac{\text{Log}_{\phi_i}(\phi_j)}{d(\phi_i,\phi_j)}
\end{equation}

Using the chain rule, the gradient of a loss function involving hyperbolic distance $h = g \circ d^p$ ($g$ being a differentiable function) can be computed as follows:

$$\nabla_{\phi_i}h=g'(d^p(\phi_i,\phi_j)) \nabla _{\phi_i} d^p(\phi_i,\phi_j)$$
where the expression of $\nabla _{\phi_i} d^p(\phi_i,\phi_j)$ is given in Equation (\ref{eq:distance_hyperbolic_gradient}).

In the attached code (discussed in subsequent sections), optimisation is performed by redefining the gradient of the distance and then using usual auto-derivation tools provided by PyTorch backend.

To avoid division by $d(\phi_i,\phi_j)$, which becomes computationally difficult when $\phi_i$ and $\phi_j$ are close, we adopt $p=2$ which experimentally revealed to be numerically more stable. Explicit forms for the gradient of $O_1$, $O_2$ and $O_3$ are as follows:

 \paragraph{Update based on $O_1$.} 
 
$$O_1 = -\sum \limits_{(v_i, v_j) \in E} log(\sigma(-d^2(\phi_i, \phi_j)))$$

Recall that $O_1$ maintains first-order proximity by preserving the distance between directly connected nodes $(v_i,v_j)\in E$ with embeddings $(\phi_i,\phi_j)$. The gradient of $O_1$ with respect to $\phi_i$ is

 $$\nabla_{\phi_i}O_1 = -2\times \text{Log}_{\phi_i}(\phi_j) \times \sigma(d^2(\phi_i,\phi_j))$$

In this last formula, we used the fact that $log(\sigma(-x))' = -\sigma(x)$. By symmetry, $\nabla_{\phi_j}O_1= -2\times \text{Log}_{\phi_j}(\phi_i) \times \sigma(d^2(\phi_i,\phi_j))$.\\

 \textbf{Update based on $O_2$.} 
$$O_2 = - \sum\limits_{v_i\in V} \sum\limits_{v_j \in C_i} \bigg[ log(\sigma(-d^2(\phi_i, \phi_j')))  + \sum \limits_{v_k\sim \mathcal{P}_n} log(\sigma(d^2(\phi_i, \phi_k')))\bigg]$$
  The updates of $\phi_i, \phi'_j$ and $\phi'_k$ the embeddings of $v_i, v_j$ and $v_k$ where $v_i$ is a node, $v_j$ belongs to the context $C_i$ of $v_i$ and $v_k$ a negative sample of $v_i$ are given as follows:\\

\hspace{0.5cm}$\bullet$ Update of $\phi'_j$ is done exactly as in $O_1$ since it occurs only in the first term of the sum.\\

\hspace{0.5cm} $\bullet$ Update of $\phi'_k$ is based on the gradient computation \\

 $$\nabla_{\phi'_k}O_2 = -2\times \text{Log}_{\phi'_k}(\phi_i) \times \sigma(-d^2(\phi_i,\phi'_k)) $$

\hspace{0.5cm} $\bullet$ Update of $\phi_i$ is based on the gradient computation\\

\begin{align*}
    \nabla_{\phi_i} O_2 = \sum_{v_j\in C_i} \bigg[-2\times \text{Log}_{\phi_i}(\phi'_j)\times \sigma(d^2(\phi_i,\phi'_j))
    +\sum_{k=1}^M \left(2\times \text{Log}_{\phi_i}(\phi'_
    k)\times \sigma(-d^2(\phi_i,\phi'_k)) \right) \bigg] 
\end{align*}
where $M$ is the number of negative samples.

\paragraph{Update based on $O_3$.}
\begin{align*}
  O_3 = - \sum \limits_{v_i\in V}\sum \limits_{k=0}^K w_{ik}log\left( \frac{1}{\zeta_m(\sigma_k)}e^{-\frac{d^2(\phi_i, \mu_k)}{2\sigma_k^2}} \right)
\end{align*}

The updates of $w_{ik}, \zeta_m(\sigma_k)$ and $\sigma_k$ are detailed in the paper. The update of $\phi_i$ uses the formula

$$ \nabla_{\phi_i}O_3 = \sum_{k=0}^K \frac{w_{ik}}{2\sigma_k^2}\nabla_{\phi_i}(d^2(\phi_i,\mu_k)) $$

\section{Datasets and additional results}

In this section, additional experiments and comparisons are provided. We propose to compare our method with the baseline of the Most Common Community (MCC, described next).
This comparison will illustrate difficulties for classification algorithms to deal with GSD exhibiting imbalanced community distributions when embedded in low dimensions.

In particular, it is the case for \textit{Wikipedia} and \textit{Blogcatalog} datasets where some communities contain a single node.
Finally, thanks to the visualisation of the predictions of the different supervised algorithms, we justify the relevance of using Gaussian distributions in some situations for classification rather than classifiers based on geodesics such as the Logistic Regression.

\subsection{Comparison with Most Common Community (MCC)}

In this section, we comment the results and discuss the reasons related to classification algorithms and datasets that led to low performances for two-dimensional representations.
In particular, these situations produce low performances in the unsupervised setting which however achieve better scores in the supervised one. 
The main intuition explaining this performance gap is the tendency of supervised classification methods to annotate all nodes with the most common community in the dataset (i.e., the community that labels the highest number of nodes). To better illustrate this claim, we propose to visualise in Table \ref{tab:most_common_communities} the Most Common Community baseline (MCC). The MCC baseline associates each node with the community withdrawn from the probability distribution of the known true labels. Formally, the probability vector of the known labels $\hat{y}$ is written :
$$
    \hat{y} = \frac{\sum \limits_{i=1}^N y_i}{\sum\limits_{i=0}^N\sum\limits_{k=0}^K y_{ik}}
$$
where $y_{ik} =1$ if the true label of node $v_i$ is $k$ and $0$ if not, and $y_i$ the labelling vector of node $v_i$ (i.e., the $k$-th component of $y_i$ is 1 if $v_i$ belongs to community $k$).

\begin{figure}[ht]
    \centering
    \begin{subfigure}{.33\textwidth}
      \centering
      \includegraphics[width=.9\linewidth]{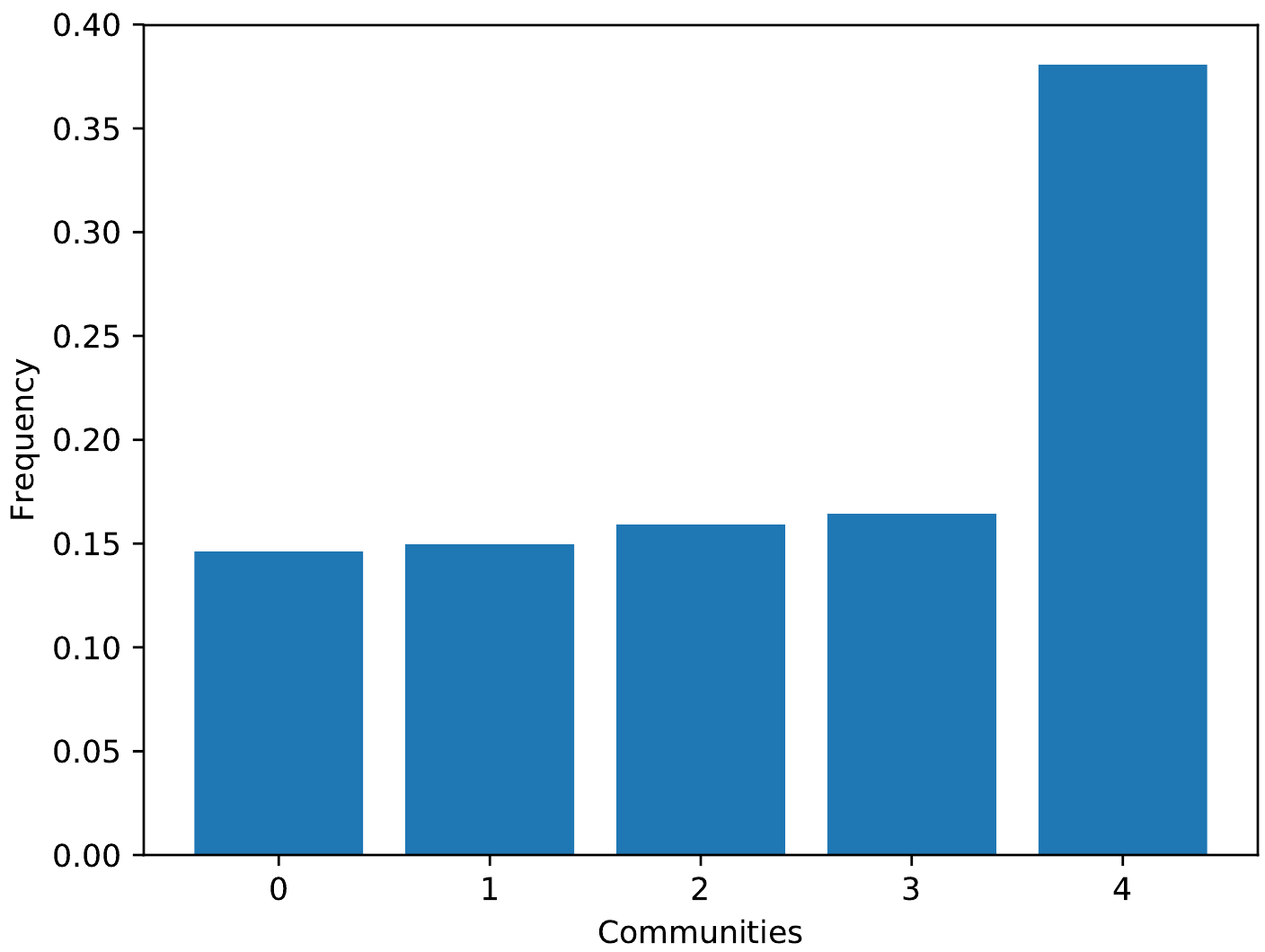}
      \caption{DBLP}
    \end{subfigure}
    \hfill
    \begin{subfigure}{.33\textwidth}
      \centering
      \includegraphics[width=.9\linewidth]{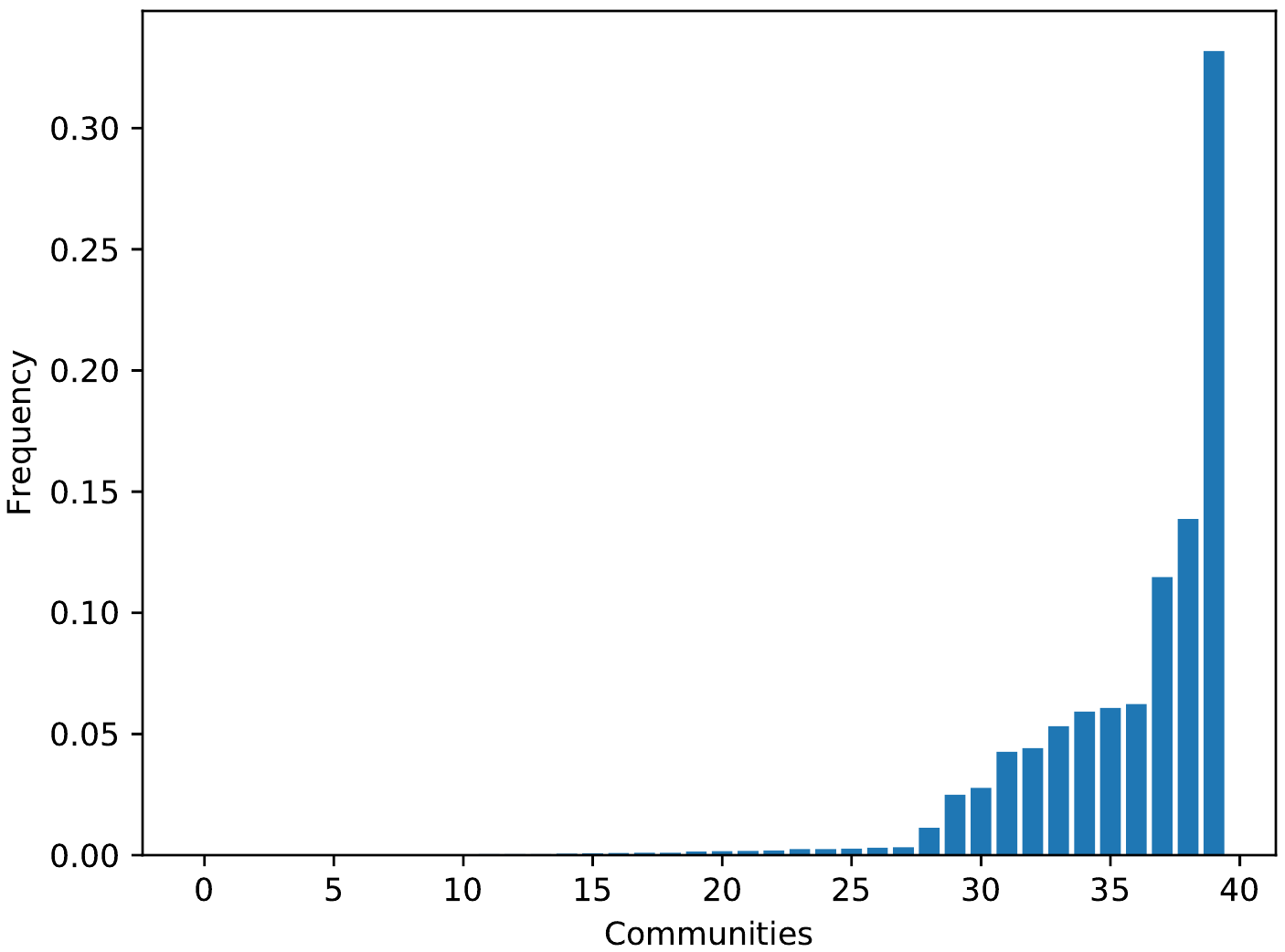}
      \caption{Wikipedia}
    \end{subfigure}
    \hfill
    \begin{subfigure}{.33\textwidth}
      \centering
      \includegraphics[width=.9\linewidth]{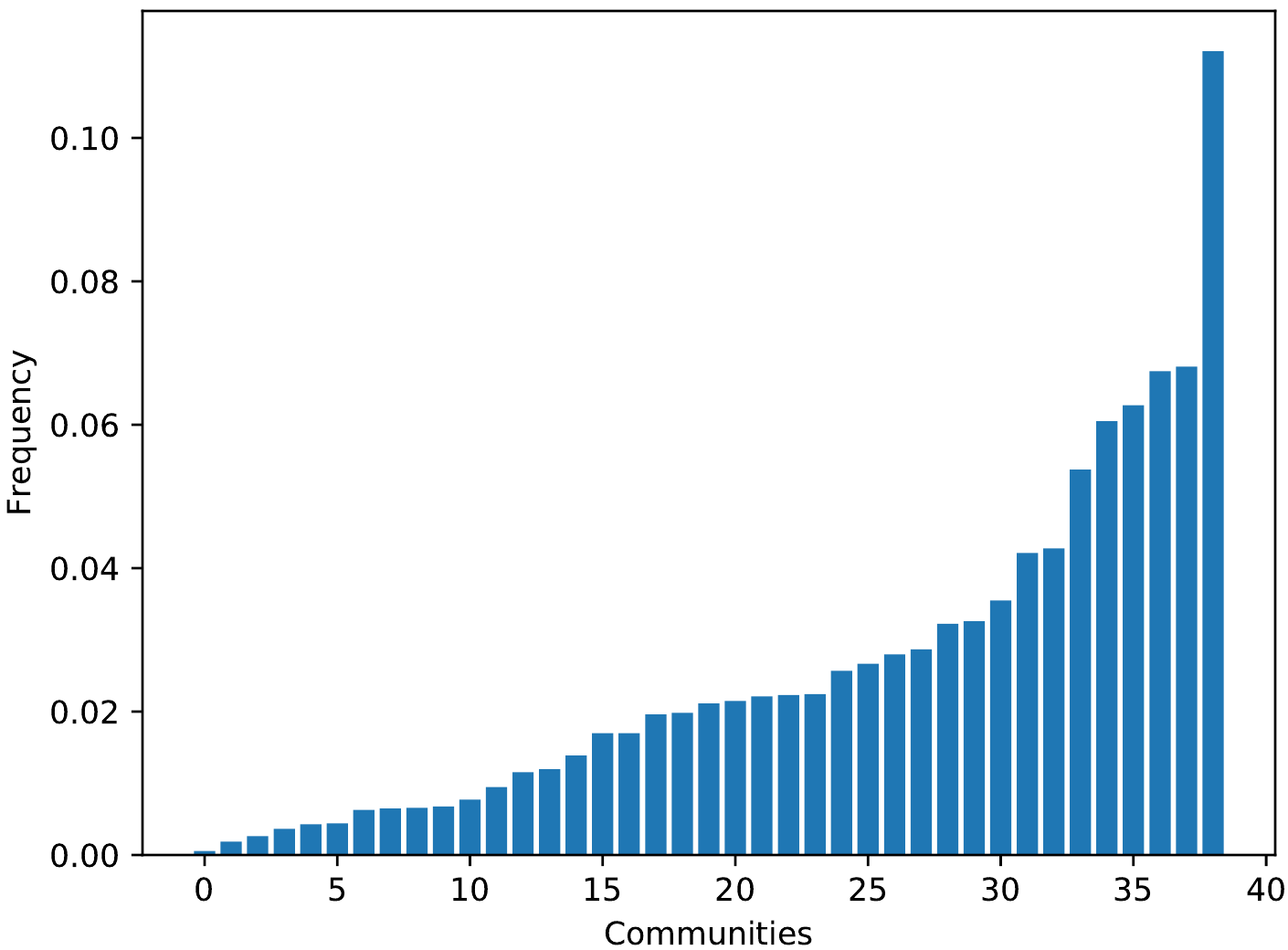}
      \caption{BlogCatalog}
    \end{subfigure}

\caption{The distributions $\hat{y}$ of communities for \textit{DBLP}, \textit{Wikipedia} and \textit{BLogCtalog} dataset}
\label{fig:dataset_distribution}
\end{figure}

\begin{table}[h]
    \caption{Precisions at  1, 3, 5 for HLR (Hyperbolic Logistic Regression) for 2 and 10 dimensional embeddings, MCC-CV (mean of most common community using 5-cross validation sets) and MCC-A (most common community for the entire dataset)}
    \label{tab:most_common_communities}
    \centering
    \begin{tabular}{l|c|c|c|c}
        \hline
        Dataset & HLR (2D) & HLR (10D) &MCC-CV &MCC-A\\
        \hline
        \hline
        Karate & 87& - &53.3& 50.0\\
        PoolBlog & 96 & - &52.0& 51.9\\
        Books & 82& - &46.6& 46.6\\
        Football & 39 & - &2.6 & 11.3\\
        DBLP &  79 & 92 &38.1& 38.0\\
        \hline
        \hline
        Wikipedia & 47/28/21 & 48/30/22 &47/28/20&47/28/20\\
        BlogCatalog & 16/12/11 & 34/20/15 &16/12/10& 16/12/10\\
        Flickr & 19/11/8& 32/19/14 &17/10/7&17/10/7\\
    \end{tabular}

\end{table}
Table \ref{tab:most_common_communities} empirically demonstrates that in 2 dimensions the HLR method is unable to efficiently capture the community structures of \textit{Wikipedia} and \textit{BlogCatalog} since it performs equally the same as MCC. 
Moreover, for \textit{Wikipedia} the most common community labels nearly half of the graph nodes as shown in Figure \ref{fig:dataset_distribution}. Taking a closer look at the same Figure, we notice how community distributions are largely unbalanced for \textit{Wikipedia} and \textit{BlogCatalog}.  
Furthermore, for \textit{Wikipedia} some communities are represented by a single node, making this dataset particularly difficult for classifiers to capture the least represented communities. On the contrary, the fact that communities of \textit{DBLP} are more uniformly distributed contributes in achieving better performances with HLR in only two dimensions.

\newpage
\subsection{Visualisation}
In this part, visualisations of the learned embeddings and prediction of the different classification algorithms are shown. Then advantages and disadvantages of each classifier are discussed.

\begin{figure*}[ht]

    \begin{subfigure}{.33\textwidth}
      \centering
      \includegraphics[width=.7\linewidth]{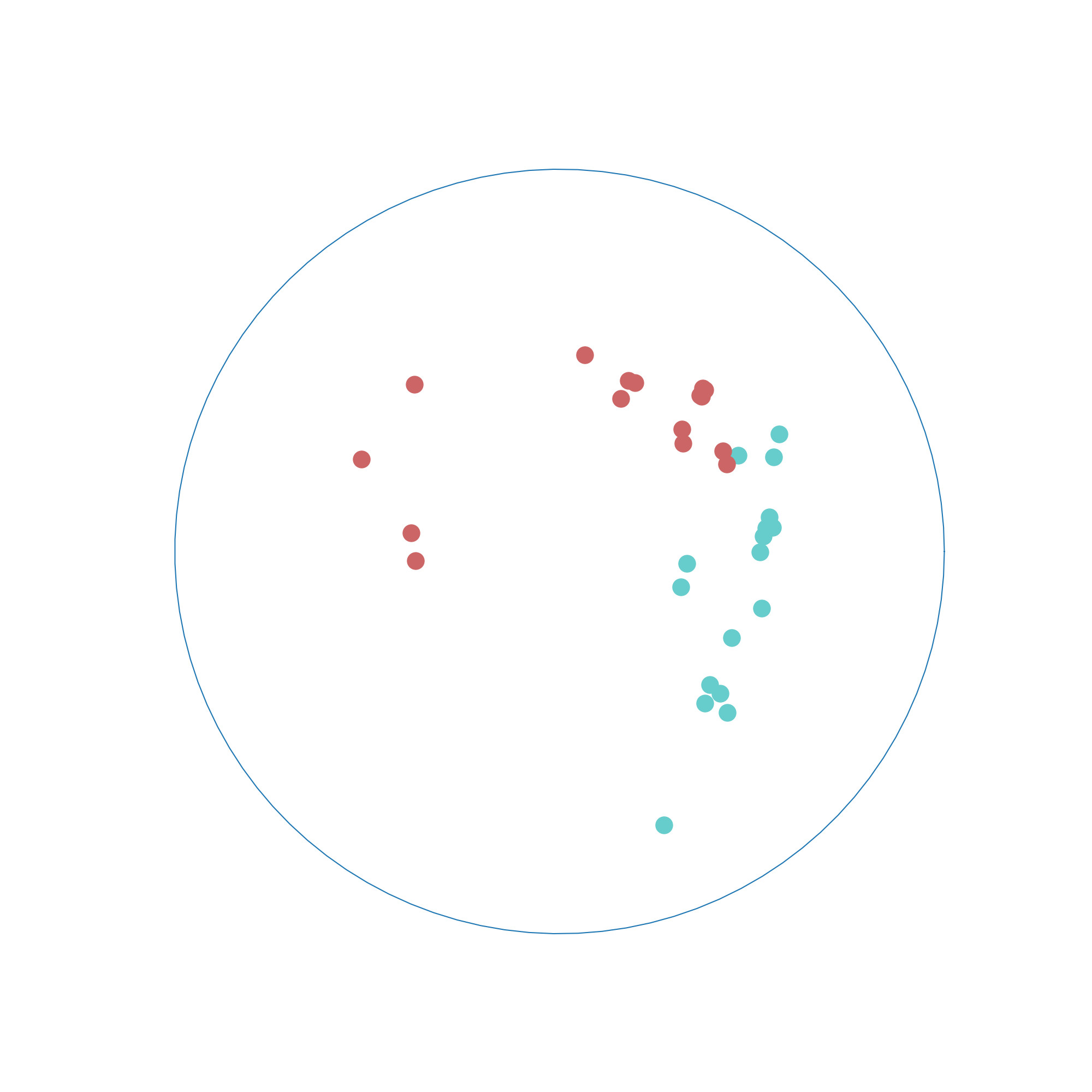}  
      \caption{Karate}
      \label{fig:sub-second}
    \end{subfigure}
    ~
    \begin{subfigure}{.33\textwidth}
      \centering
      \includegraphics[width=.7\linewidth]{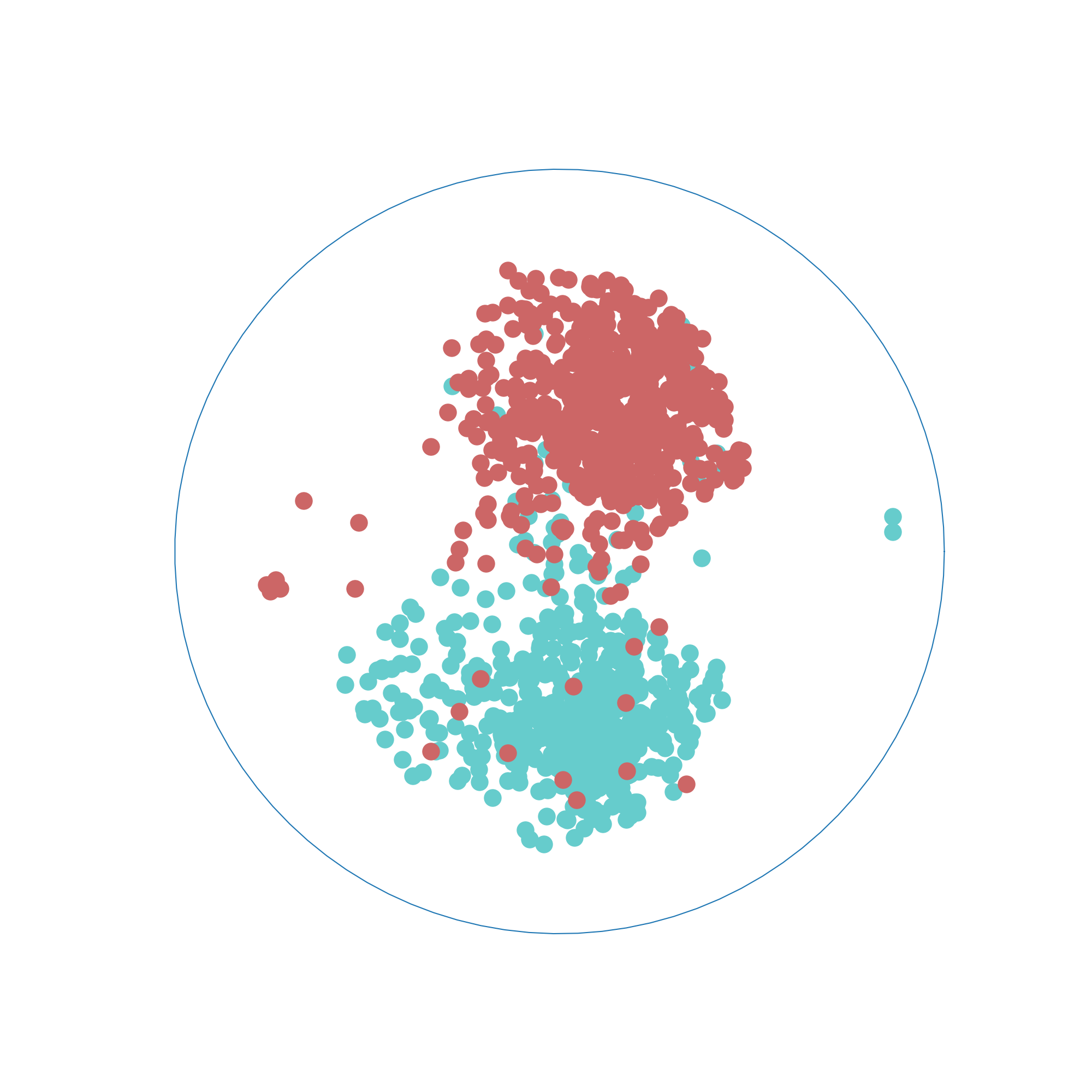}  
      \caption{PoolBlog}
      \label{fig:sub-second}
    \end{subfigure}
    ~
    \begin{subfigure}{.33\textwidth}
      \centering
      \includegraphics[width=.7\linewidth]{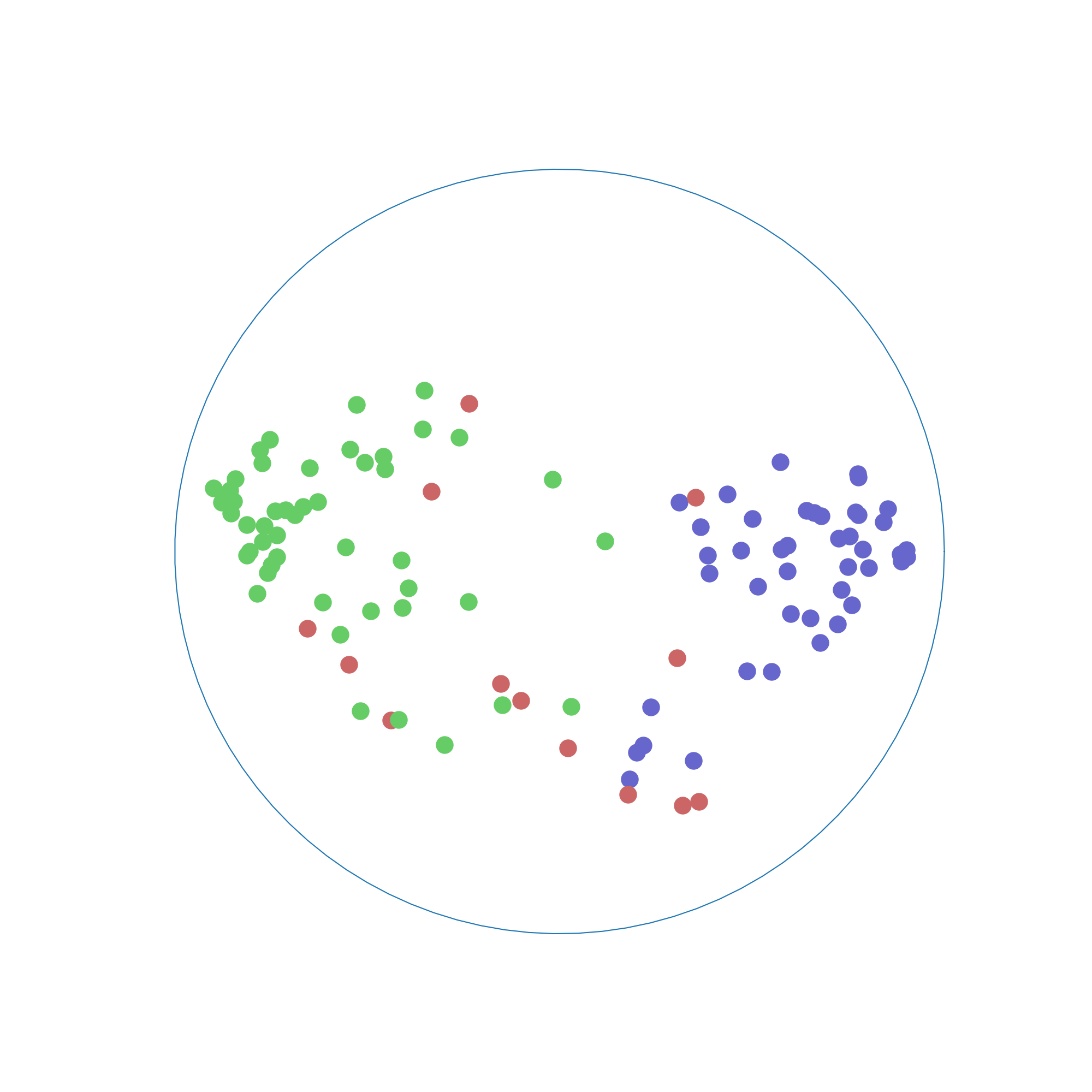}  
      \caption{Books}
      \label{fig:sub-second}
    \end{subfigure}
    
    \begin{subfigure}{.33\textwidth}
      \centering
      \includegraphics[width=.7\linewidth]{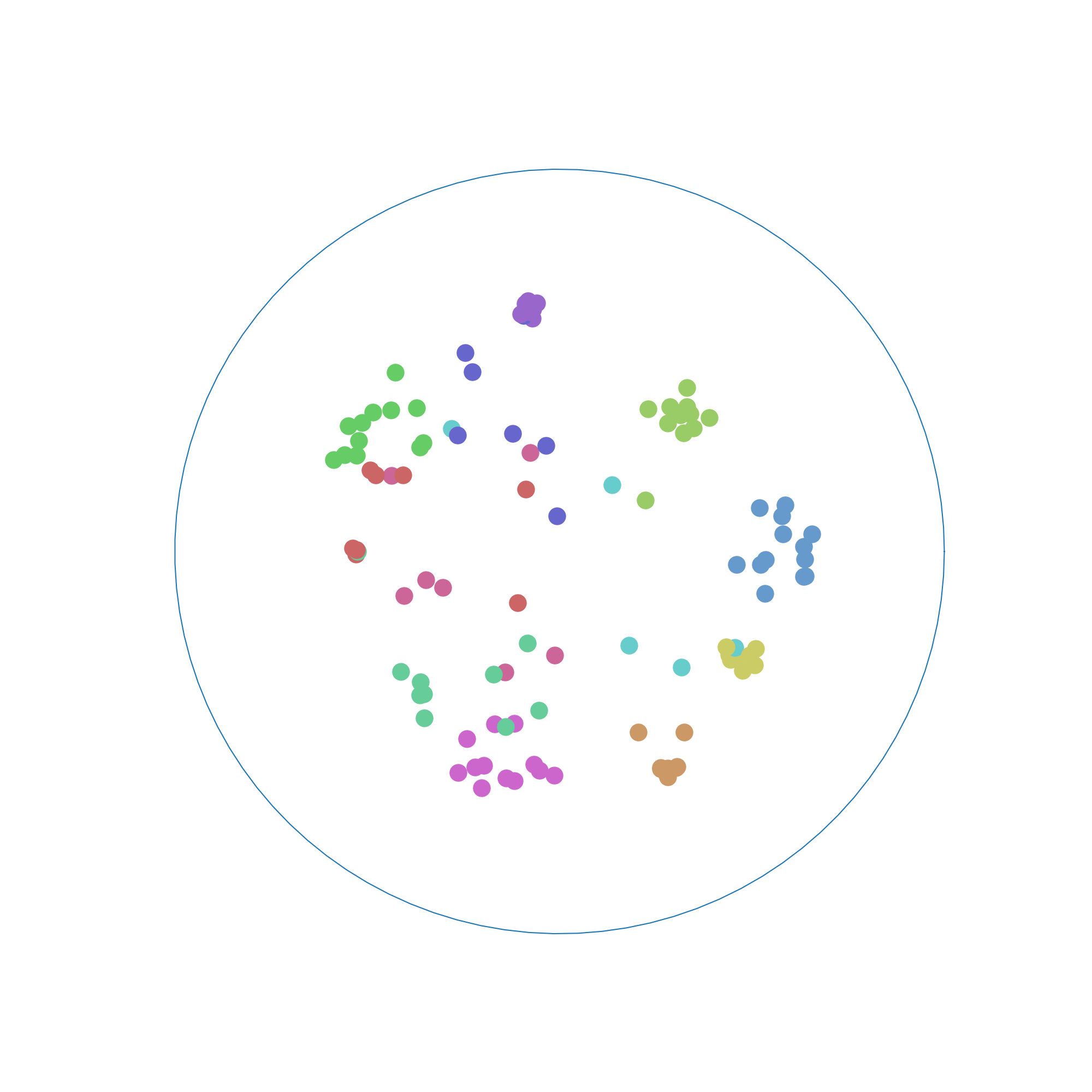}  
      \caption{Football}
      \label{fig:sub-second}
    \end{subfigure}
    ~
    \begin{subfigure}{.33\textwidth}
      \centering

      \includegraphics[width=.7\linewidth]{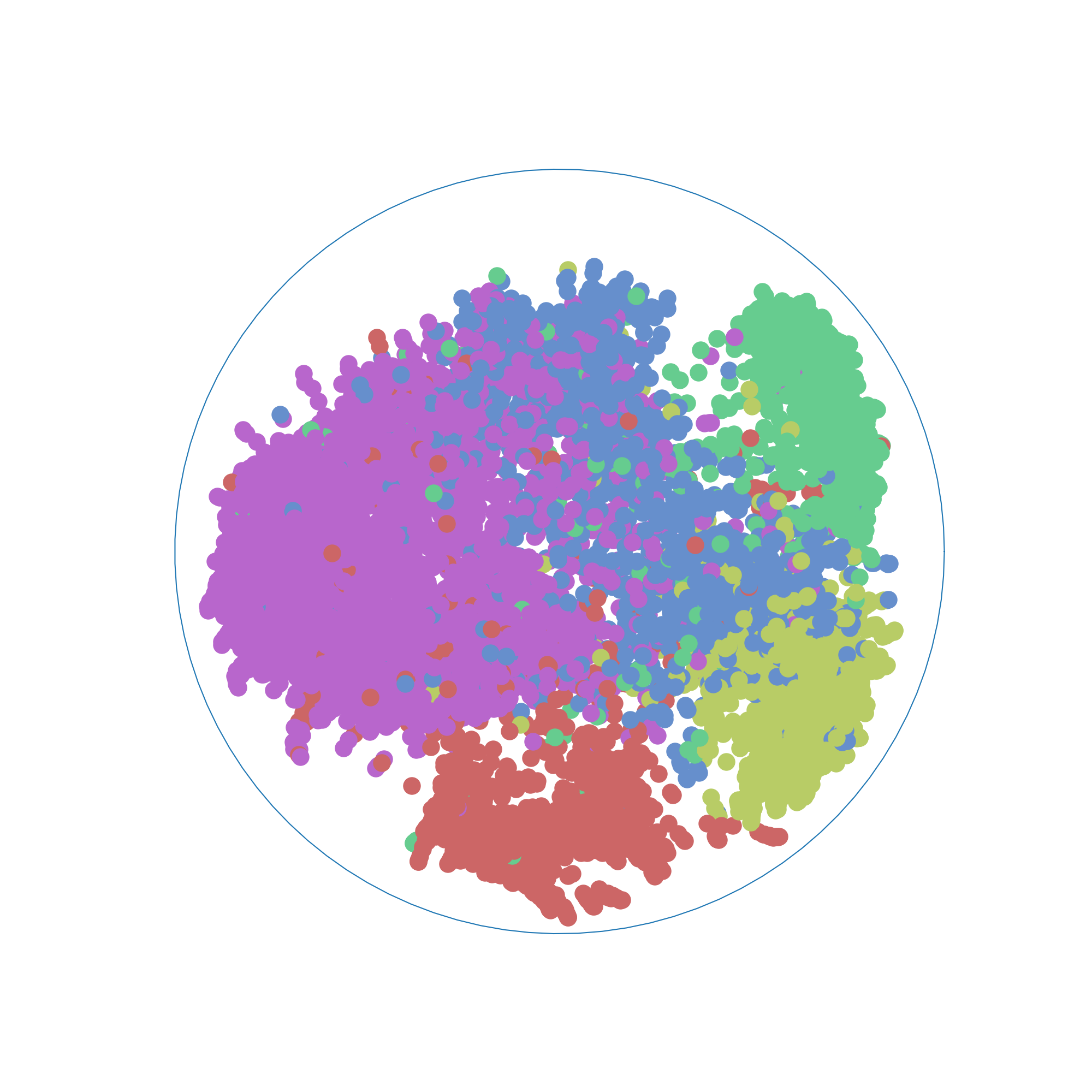}  
      \label{fig:sub-first}
      \caption{DBLP}
    \end{subfigure}

\caption{Embeddings visualisations, colors represent ground truth communities}
\label{fig:ground_truth_supplementary}
\end{figure*}

Figure \ref{fig:ground_truth_supplementary} shows the learned embeddings for the graph datasets \textit{Karate}, \textit{PoolBlog}, \textit{Books}, \textit{Football} and \textit{DBLP} where the color of each node corresponds to its real community. Notice how the embeddings lead to clear separate clusters, allowing for classification algorithms to efficiently retrieve communities.

\begin{figure*}[ht]

    \begin{subfigure}{.33\textwidth}
      \centering
      \includegraphics[width=.7\linewidth]{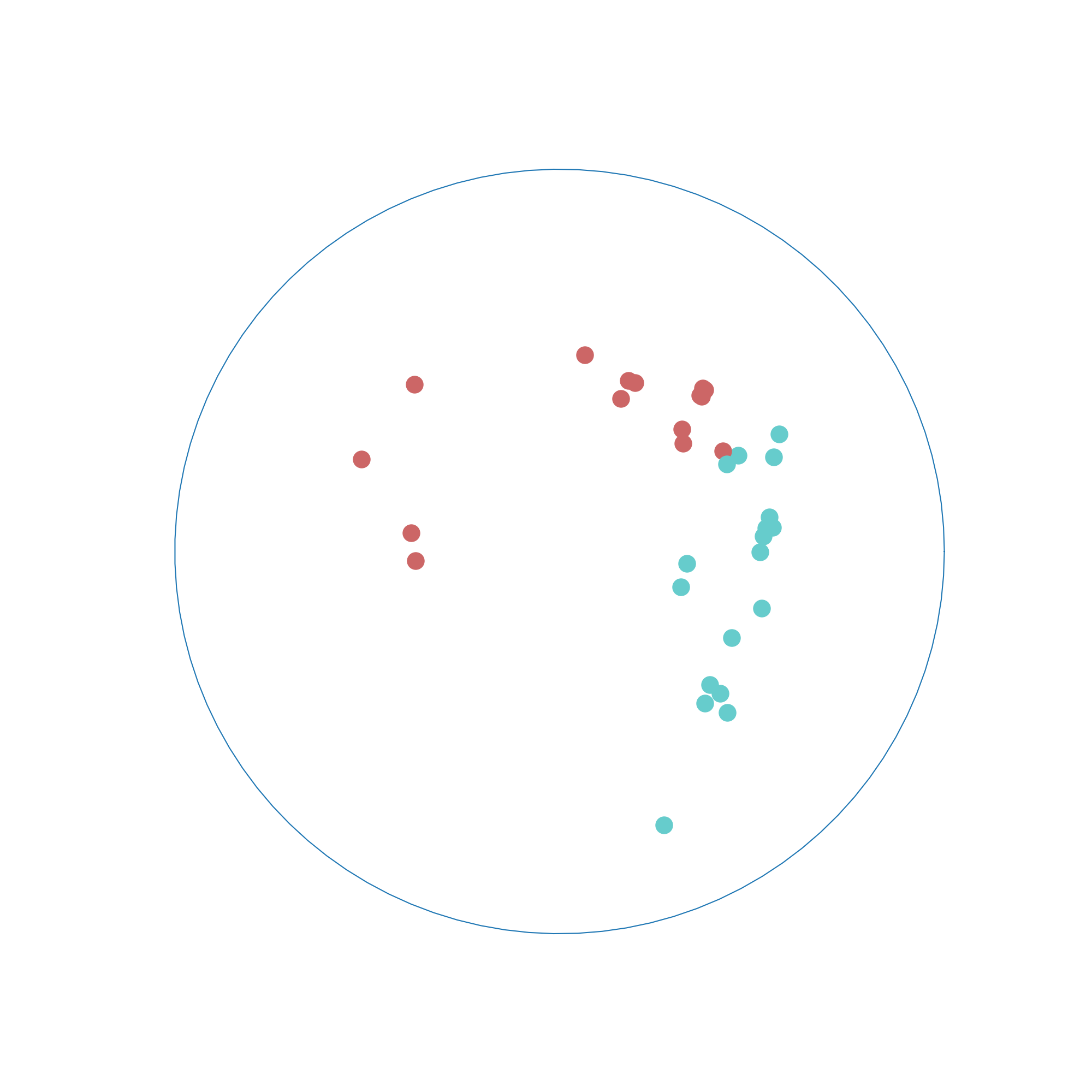}  
      \caption{Karate}
      \label{fig:sub-second}
    \end{subfigure}
    ~
    \begin{subfigure}{.33\textwidth}
      \centering
      \includegraphics[width=.7\linewidth]{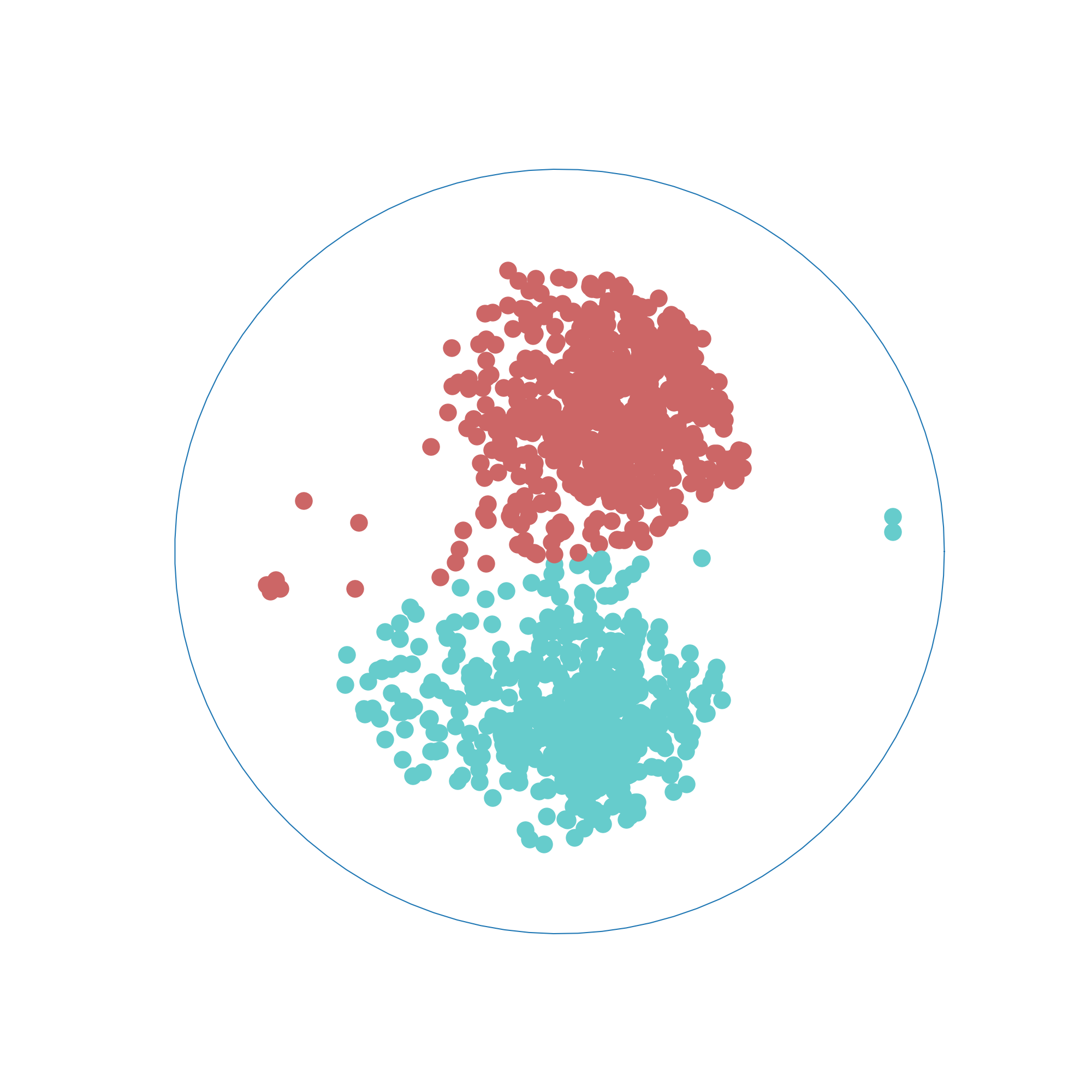}  
      \caption{PoolBlog}
      \label{fig:sub-second}
    \end{subfigure}
    ~
    \begin{subfigure}{.33\textwidth}
      \centering
      \includegraphics[width=.7\linewidth]{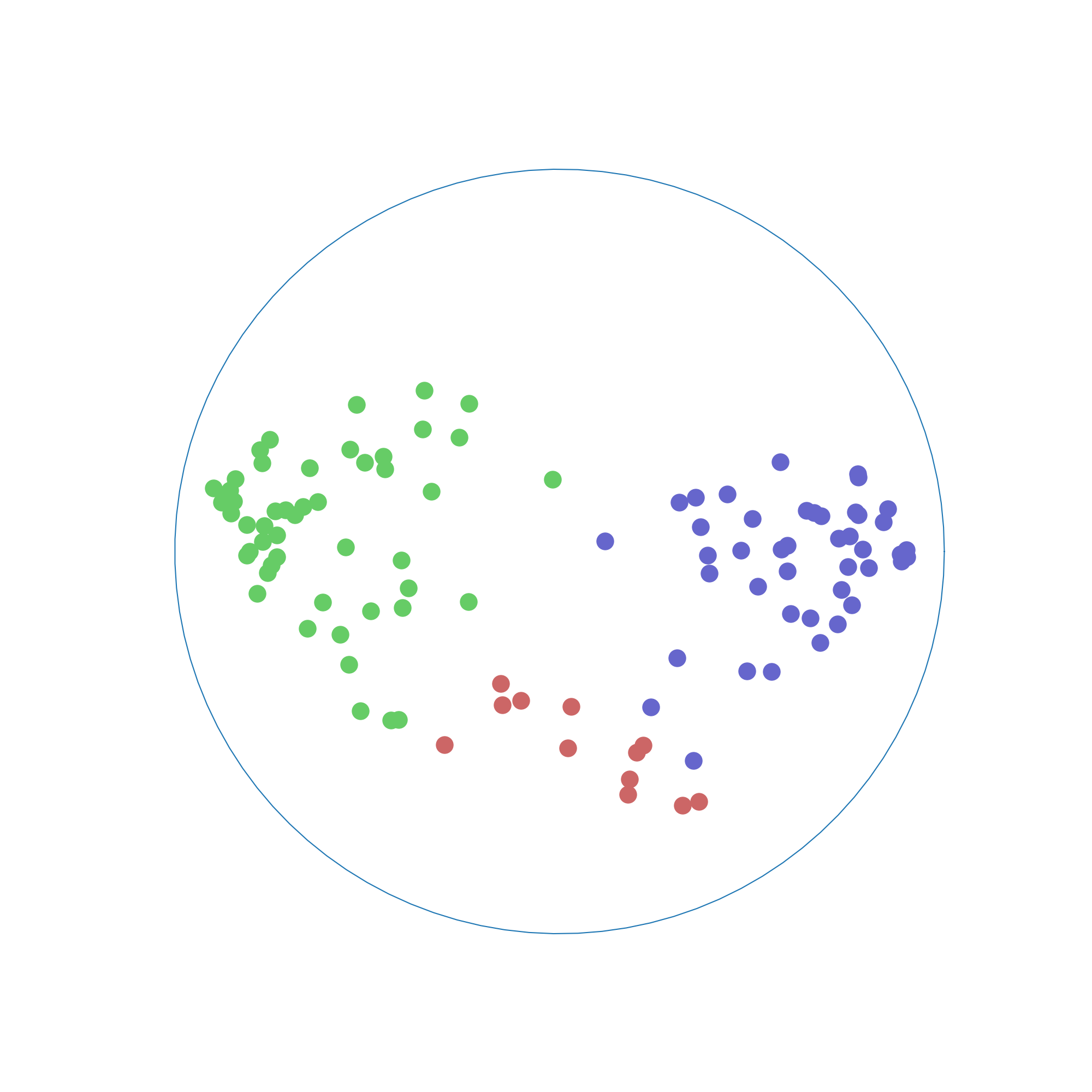}  
      \caption{Books}
      \label{fig:sub-second}
    \end{subfigure}
    
    \begin{subfigure}{.33\textwidth}
      \centering
      \includegraphics[width=.7\linewidth]{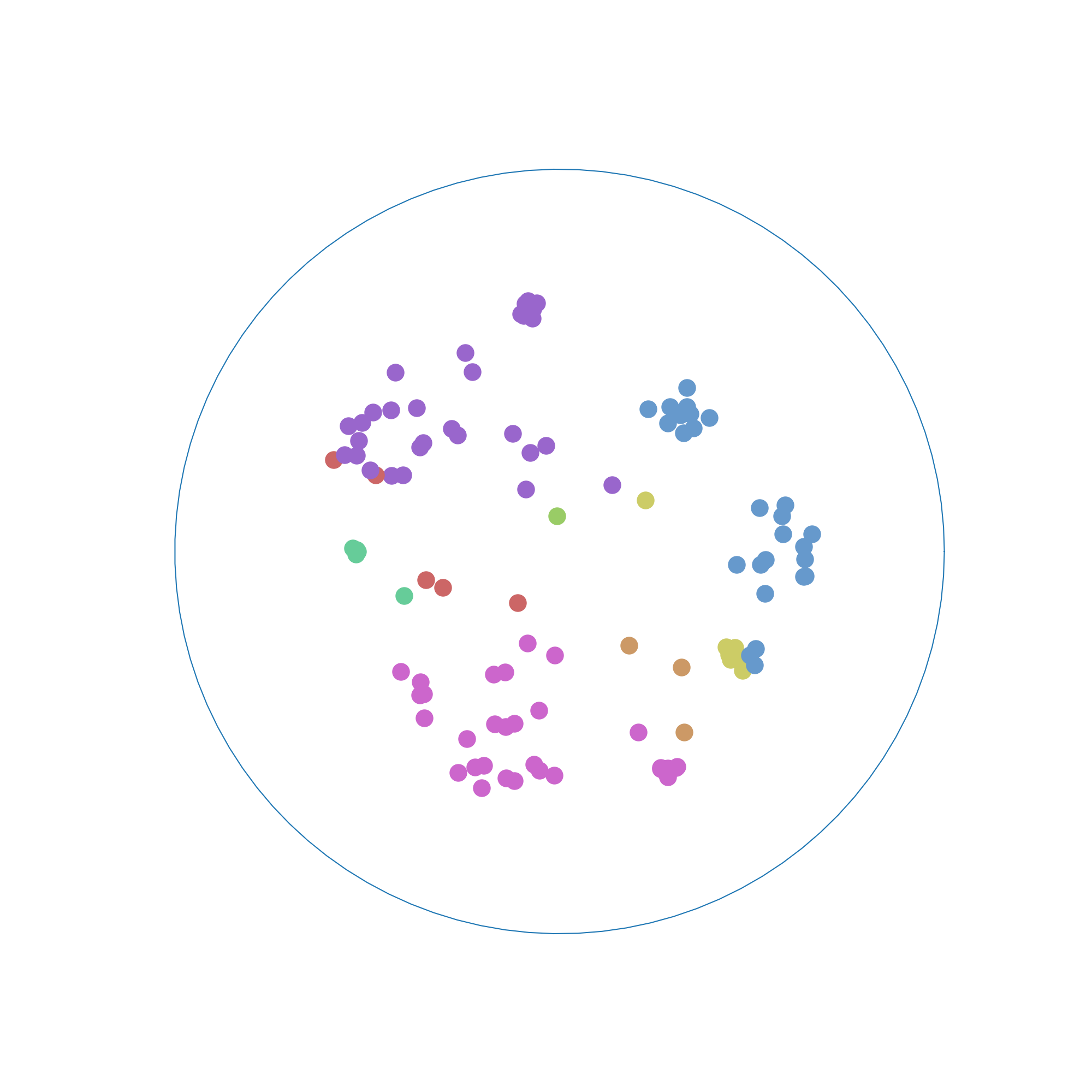}  
      \caption{Football}
      \label{fig:sub-second}
    \end{subfigure}
    ~
    \begin{subfigure}{.33\textwidth}
      \centering

      \includegraphics[width=.7\linewidth]{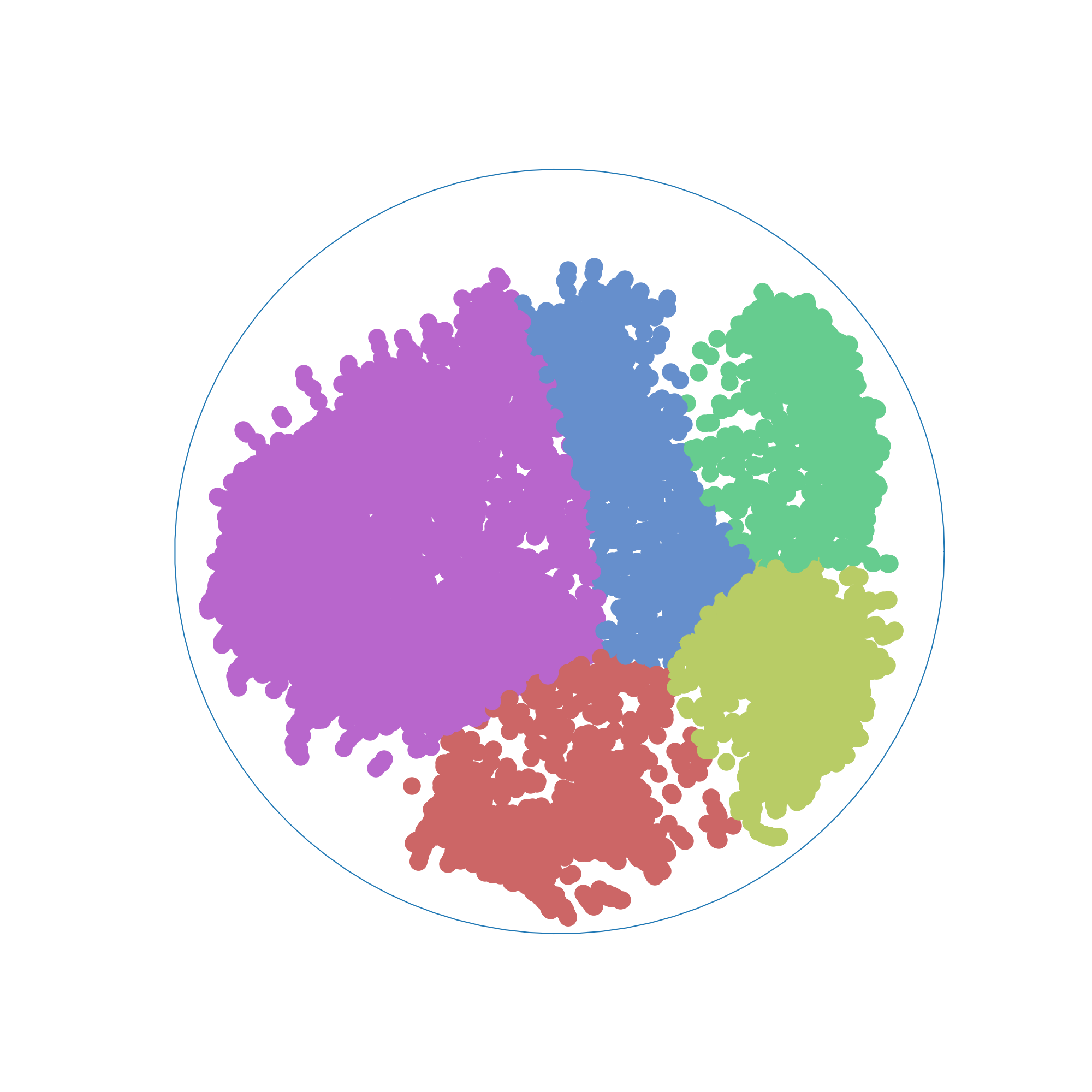}  
      \label{fig:sub-first}
      \caption{DBLP}
    \end{subfigure}

\caption{Prediction obtained using HLR classification (best view in color). Colors represent community prediction.}
\label{fig:HLR_figure}
\end{figure*}

\begin{figure*}[ht]

    \begin{subfigure}{.33\textwidth}
      \centering
      \includegraphics[width=.7\linewidth]{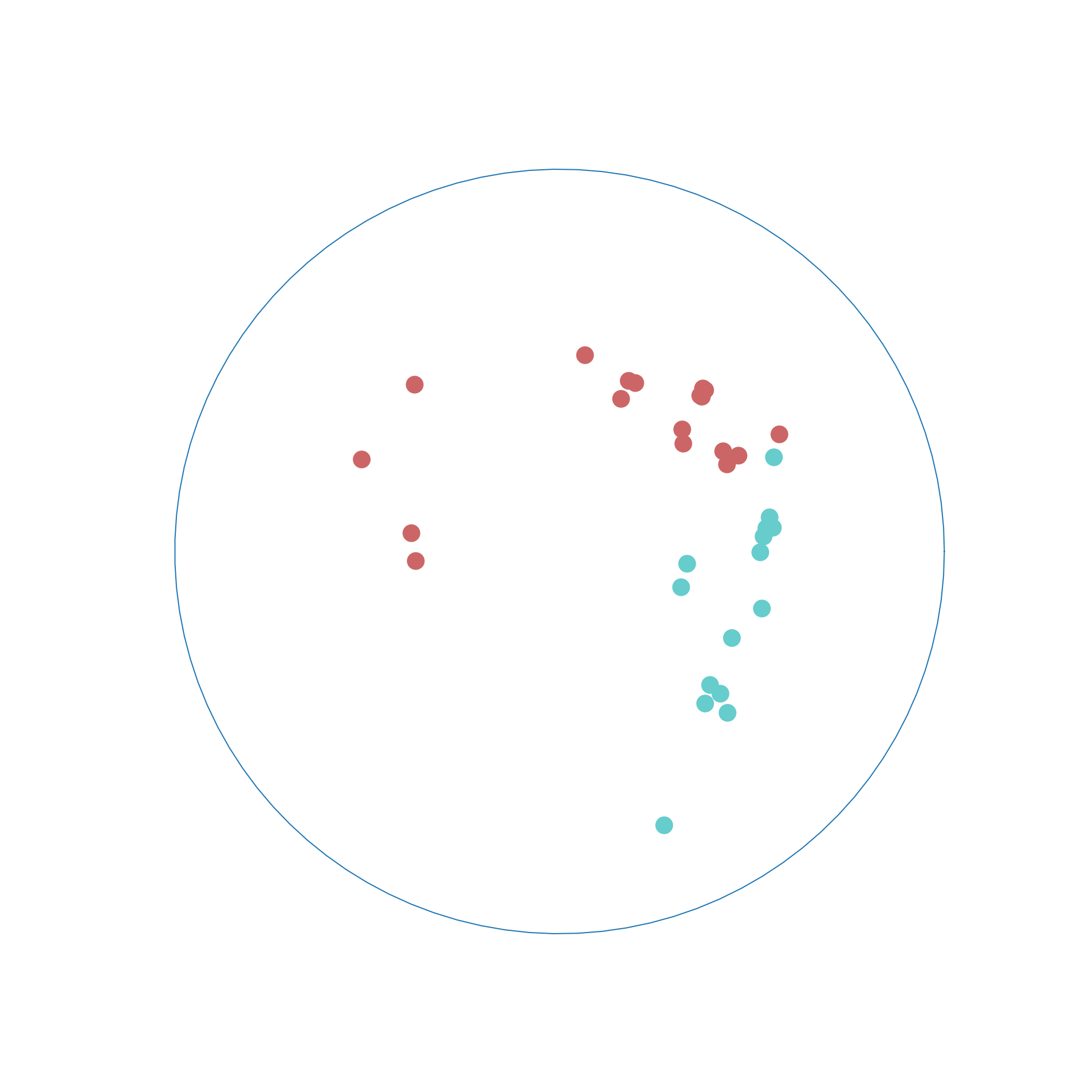}  
      \caption{Karate}
      \label{fig:sub-second}
    \end{subfigure}
    ~
    \begin{subfigure}{.33\textwidth}
      \centering
      \includegraphics[width=.7\linewidth]{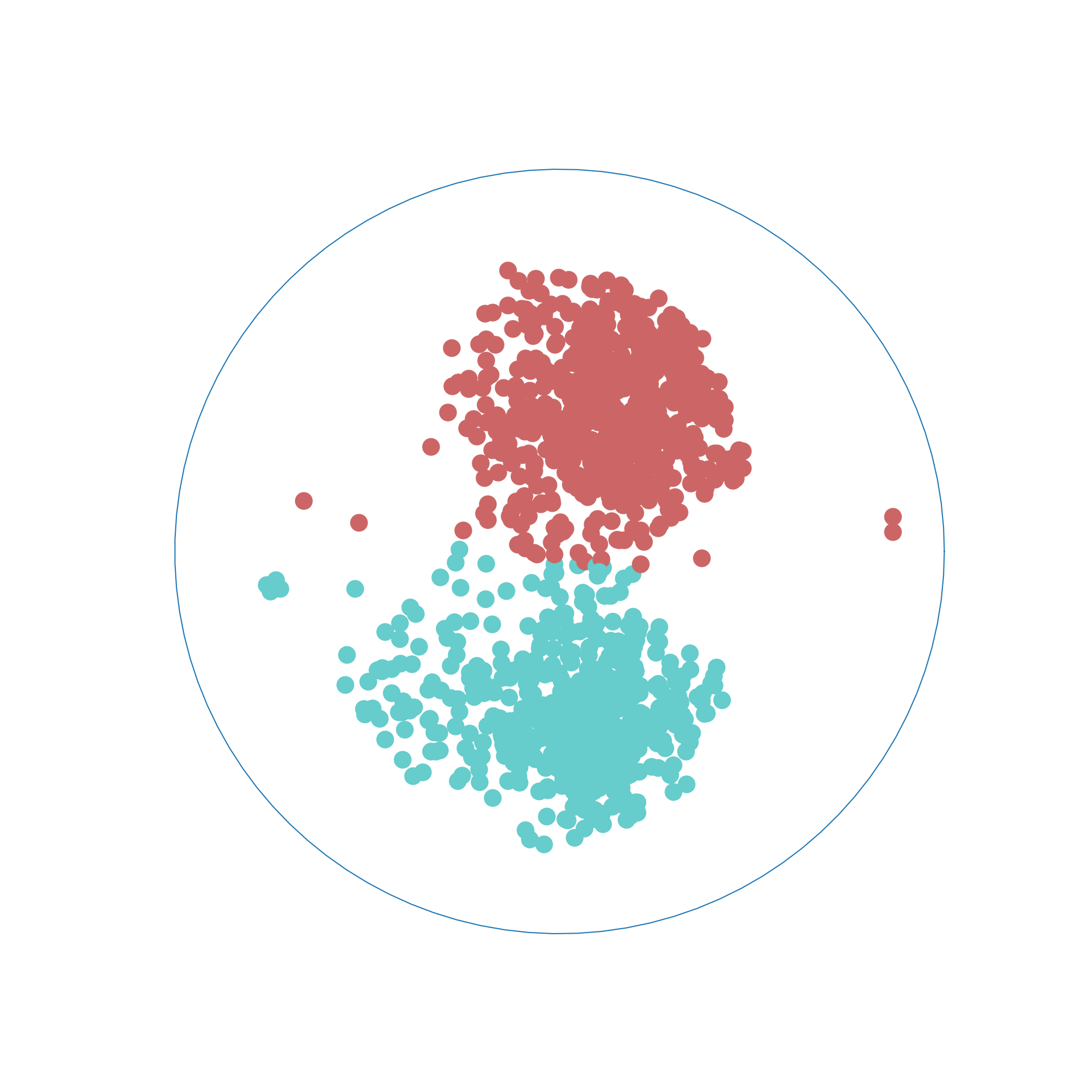}  
      \caption{PoolBlog}
      \label{fig:sub-second}
    \end{subfigure}
    ~
    \begin{subfigure}{.33\textwidth}
      \centering
      \includegraphics[width=.7\linewidth]{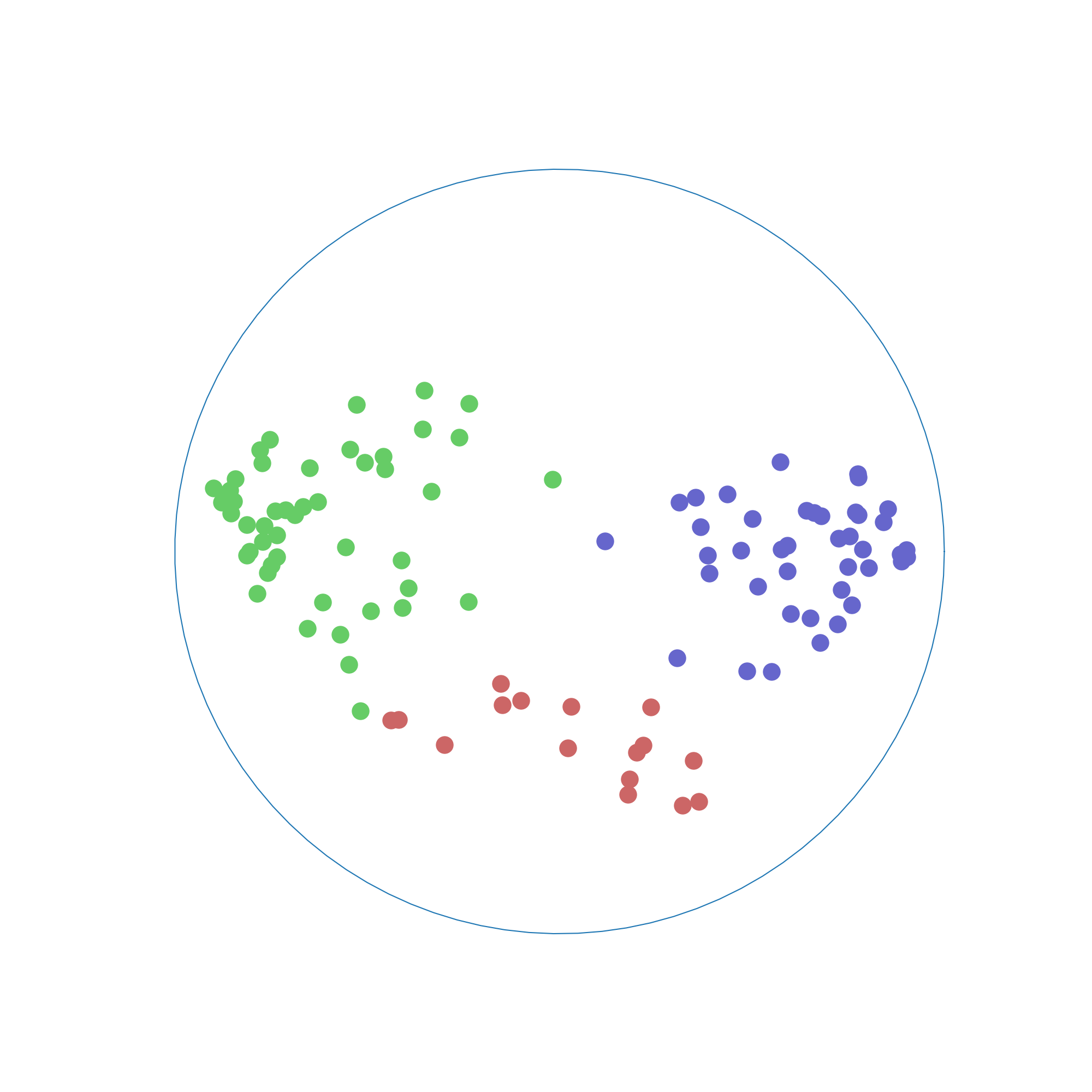}  
      \caption{Books}
      \label{fig:sub-second}
    \end{subfigure}
    
    \begin{subfigure}{.33\textwidth}
      \centering
      \includegraphics[width=.7\linewidth]{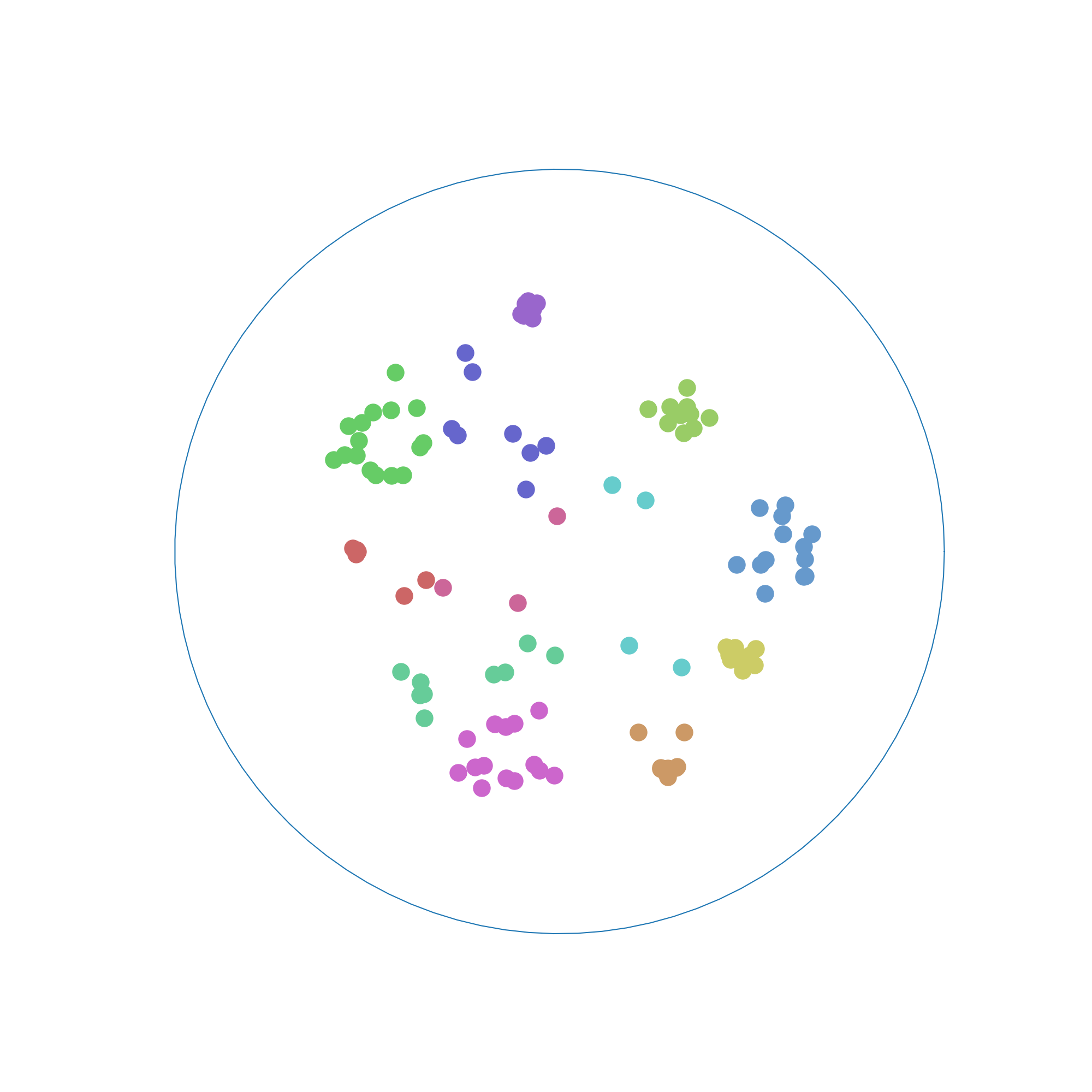}  
      \caption{Football}
      \label{fig:sub-second}
    \end{subfigure}
    ~
    \begin{subfigure}{.33\textwidth}
      \centering

      \includegraphics[width=.7\linewidth]{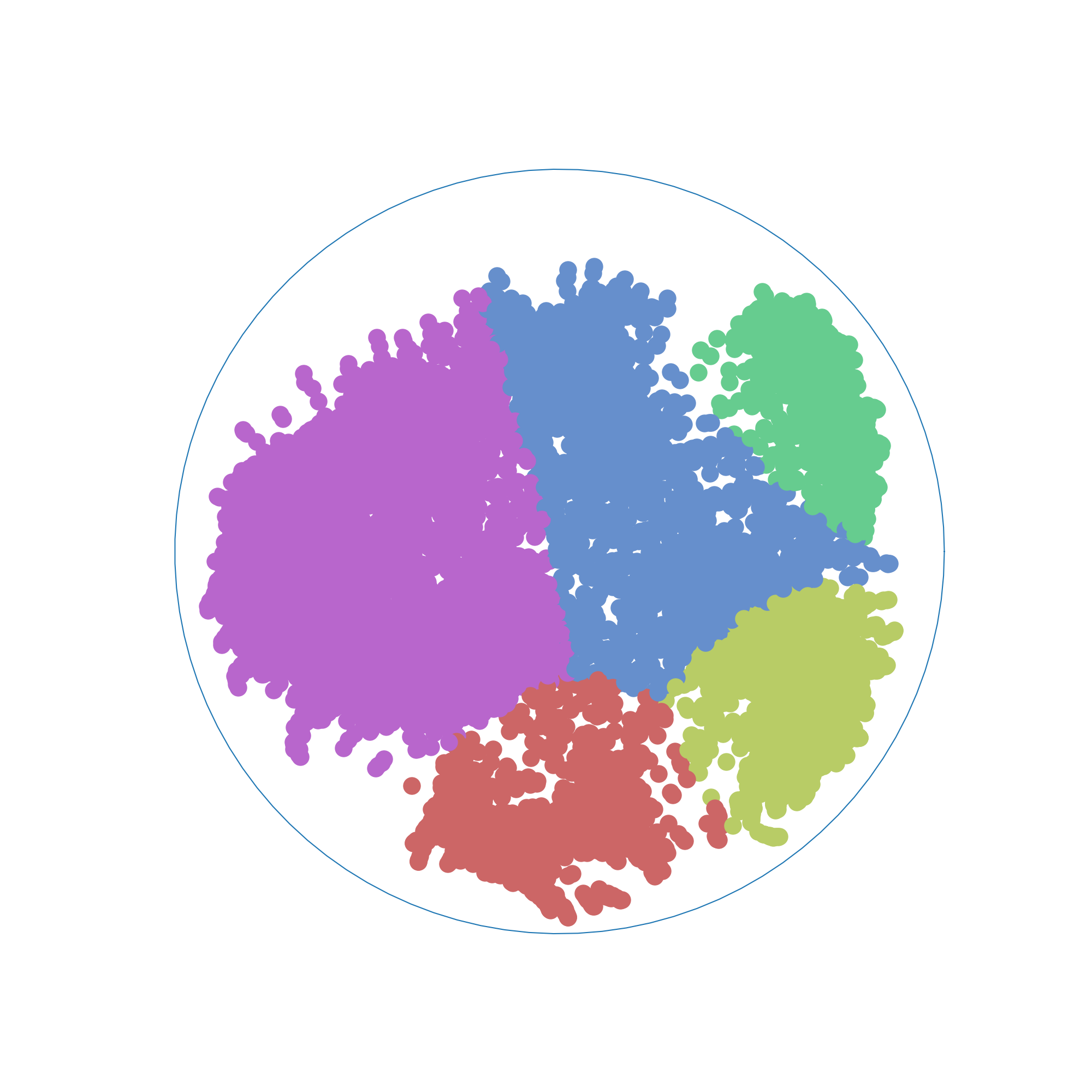}  
      \label{fig:sub-first}
      \caption{DBLP}
    \end{subfigure}

\caption{Prediction obtained using \textit{GMM} classification method (best view in color). Colors represent community prediction.}
\label{fig:gmm_figure}
\end{figure*}

However, when visually comparing the results of \textit{GMM} and \textit{HLR} in Figures \ref{fig:HLR_figure} and \ref{fig:gmm_figure}, notice how \textit{HLR} struggles to find geodesics separating accurately communities of \textit{Football}. Indeed, the learned two-dimensional embeddings are non-separable using geodesics for \textit{Football}. This difficulty is however better handled by \textit{GMM}. Moreover, \textit{HLR} performances on \textit{Football} achieve $\approx39\%$ while reaching $\approx80\%$ with \textit{GMM}, thus supporting the benefit of making use of classification with Gaussian distributions.

\clearpage
\section{Implementation}
A python code (soon to be released to the public) performs graph embedding in the Poincaré Ball $\mathbb{B}^m$ for all dimensions $m\leq 10$ and applies Riemannian versions of Expectation-Maximisation (EM) and K-means algorithm as detailed in the paper. The \texttt{\textbf{Readme}} in the code folder details the required dependencies to operate the code, how to reproduce the results of the paper as well as effective ways to run experiments (produce a grid search, evaluate performances and so on).
The code uses PyTorch backend with 64-bits floating-point precision (set by default) to learn embeddings. In this section, further details of the procedures are presented in addition to those given in the paper and the \texttt{\textbf{Readme}}.

\subsection{EM Algorithm}

\begin{itemize}
  \item \textbf{Weighted Barycenter:} We set for variable $\lambda$ (learning rate) and $\epsilon$ (convergence rate) in Algorithm 1 of the paper the values $\lambda=5e-2$ and $\epsilon=1e-4$.
  \item \textbf{Normalisation coefficient :} We compute the normalisation factor of the Gaussian distribution for $\sigma$ in the interval [1e-3,2] with step size of 1e-3. This is a quite important parameter since if the minimum value of sigma is too high then unsupervised precision is better on datasets for which it is difficult to separate clusters in small dimensions (\textit{Wikipedia}, \textit{BlogCatalog} mainly) as discussed in the previous MCC section (for \textit{Wikipedia} most common community labelled $\approx47\%$ of the nodes and $\approx17\%$ for \textit{BlogCatalog}).
  \item \textbf{EM convergence :} In the provided implementation, the EM is considered to have converged when the values of $w_{ik}$ change less than 1e-4 w.r.t the previous iteration, more formally when:
    $$
      \frac{1}{N}\sum\limits_{i=0}^N\frac{1}{K}\sum\limits_{k=0}^K(|w_{ik}^t-w_{ik}^{t+1}|)<1e-4
    $$
  For instance using \textit{Flickr} the first update of \textit{GMM} distribution converged in approximately 50 to 100 iterations.
  
\end{itemize}

\subsection{Learning embeddings}

\begin{itemize}
  \item \textbf{Optimisation :} In some cases due to the Poincaré ball distance, the updates of the form $\text{Exp}_u(\eta\nabla_uf(d(u,v))$ can reach a norm of 1 (because of floating point precision). In this special case we do not take into account the current gradient update. If it occurred too frequently we recommend to lower the learning rate.
  \item \textbf{Moving context size :} Similarly to \textit{ComE}, we use a moving size window on the context instead of a fixed one thus we uniformly sample the size of the window between the max size given  as input argument and one.
\end{itemize}

\end{document}